\pdfoutput=1
\documentclass[10pt, logo, onecolumn, copyright]{nv}%
\usepackage{graphicx}
\definecolor{nvidiagreen}{HTML}{76B900}

\usepackage{mdframed}
\usepackage{color}
\usepackage{xcolor}
\usepackage[table]{xcolor}
\usepackage[utf8]{inputenc}
\usepackage[T1]{fontenc}

\usepackage{amsfonts} 
\usepackage{nicefrac} 
\usepackage{microtype}
\usepackage{multirow}
\usepackage{multicol}
\usepackage{tabto}
\usepackage{xspace}
\usepackage{amsmath}
\usepackage{adjustbox}
\usepackage{enumitem}
\usepackage{wrapfig}
\usepackage{dblfloatfix}
\usepackage{times}
\usepackage{verbatim}
\usepackage{amssymb}
\usepackage{mathtools}
\usepackage{caption}
\usepackage{subcaption}
\usepackage{array}
\usepackage{colortbl}
\usepackage{booktabs}
\usepackage{bbm}
\usepackage{makecell}
\usepackage{float}
\usepackage{siunitx}
\usepackage{pifont}
\usepackage{marvosym}
\usepackage{listings}
\usepackage{pdflscape}
\usepackage{footmisc}
\usepackage{url}
\usepackage{tabularx}
\usepackage{arydshln}
\usepackage{hhline}
\usepackage{diagbox}
\usepackage{tcolorbox}
\usepackage[nameinlink]{cleveref}
\usepackage{hyperref}
\usepackage[square,sort,comma,numbers]{natbib} 
\usepackage{fp} 
\usepackage{authblk}
\usepackage{xspace}

\crefname{section}{Sec.}{Sec.}
\crefname{proposition}{Proposition.}{Proposition.}
\crefname{equation}{Eq.}{Eqs.}
\crefname{figure}{Fig.}{Figs.}
\crefname{table}{Tab.}{Tabs.}
\crefname{algorithm}{Algorithm}{Algorithms}
\crefname{appendix}{Appendix}{Appendices}
\Crefname{thm}{Thm}{Thm}
\newcommand{\xmarker}{\ding{55}}
\newcommand{\cmarker}{\ding{51}}

\definecolor{codebg}{RGB}{245, 245, 245} 
\definecolor{keywordcolor}{RGB}{0, 0, 153} 
\definecolor{commentcolor}{RGB}{34, 139, 34} 
\definecolor{stringcolor}{RGB}{163, 21, 21}
\definecolor{numbercolor}{RGB}{128, 128, 128}

\title{FP4 Explore, BF16 Train: Diffusion Reinforcement Learning via Efficient Rollout Scaling}

\author{
\parbox{\linewidth}{
\centering
\vspace{-5pt}
\fontsize{9.5pt}{18pt}\selectfont
Yitong Li\textsuperscript{1,2$*$}, 
Junsong Chen\textsuperscript{1,2$*$}, 
Shuchen Xue\textsuperscript{1$*$}, 
Pengcuo Zeren\textsuperscript{1}, 
Siyuan Fu\textsuperscript{1}, 
Dinghao Yang\textsuperscript{1},
Yangyang Tang\textsuperscript{1}, 
Junjie Bai\textsuperscript{1}, 
Ping Luo\textsuperscript{2}, 
Song Han\textsuperscript{1,3}, 
Enze Xie\textsuperscript{1} \\ 
\vspace{2mm}

{\normalsize \textsuperscript{1}NVIDIA, \textsuperscript{2}HKU, \textsuperscript{3}MIT} \\
\vspace{2pt}
{\footnotesize $^*$Equal contribution.}
} 
}

\begin{abstract}
\vspace{-10pt}

Reinforcement-Learning-based post-training has recently emerged as a promising paradigm for aligning text-to-image diffusion models with human preferences. In recent studies, increasing the rollout group size yields pronounced performance improvements, indicating substantial room for further alignment gains. However, scaling rollouts on large-scale foundational diffusion models (e.g., FLUX.1-12B) imposes a heavy computational burden. To alleviate this bottleneck, we explore the integration of FP4 quantization into Diffusion RL rollouts. Yet, we identify that naive quantized pipelines inherently introduce risks of performance degradation. To overcome this dilemma between efficiency and training integrity, we propose \textbf{Sol-RL} (Speed-of-light RL), a novel \textbf{FP4-empowered Two-stage Reinforcement Learning} framework.
First, we utilize high-throughput NVFP4 rollouts to generate a massive candidate pool and extract a highly contrastive subset. Second, we regenerate these selected samples in BF16 precision and optimize the policy exclusively on them. By decoupling candidate exploration from policy optimization, Sol-RL integrates the algorithmic mechanisms of rollout scaling with the system-level throughput gains of NVFP4. This synergistic algorithm-hardware design effectively accelerates the rollout phase while reserving high-fidelity samples for optimization. We empirically demonstrate that our framework maintains the training integrity of BF16 precision pipeline while fully exploiting the throughput gains enabled by FP4 arithmetic. Extensive experiments across SANA, FLUX.1, 
and SD3.5-L substantiate that our approach delivers superior alignment performance across multiple metrics while accelerating training convergence by up to \textbf{$4.64\times$}, 
unlocking the power of massive rollout scaling at a fraction of the cost.
    \newline
    \textbf{Links:} \hspace{2pt}
    {\hypersetup{urlcolor=nvidiagreen}
    \href{https://github.com/NVlabs/Sana/}{Github Code} |
    \href{https://nvlabs.github.io/Sana/Sol-RL/} {Project Page}
    }
\end{abstract}

\begin{document}

\maketitle

\vspace{-10pt}
\begin{figure}[h]
\centering
\includegraphics[width=0.98\textwidth]{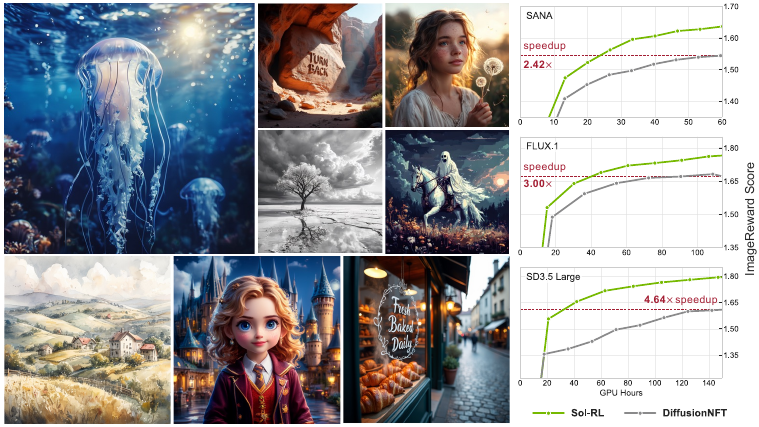}
\caption{\textbf{Sol-RL enables efficient and high-fidelity text-to-image alignment.} (Left) High-quality images generated by FLUX.1 and SANA fine-tuned with our method, demonstrating superior generation capabilities across diverse styles. (Right) ImageReward training curves. They demonstrate that Sol-RL achieves substantial wall-clock speedups (up to $\mathbf{4.64\times}$) to reach an equivalent reward level, ultimately converging to a higher alignment ceiling.}
\label{fig:teaser}
\end{figure}

\section{Introduction}
Reinforcement Learning (RL) has emerged as a highly effective paradigm for aligning Large Language Models
(LLMs) with human preferences~\citep{schulman2017proximal, bai2022training, ouyang2022training}. Particularly, Group Relative Policy Optimization (GRPO)~\citep{shao2024deepseekmath, guo2025deepseek} offers a scalable, critic-free alternative that considerably reduces training overhead while maintaining highly competitive alignment capabilities.
Building upon this success, recent advancements~\citep{liu2025flow,xue2025dancegrpo,diffusionnft,xue2025advantage} have adapted GRPO to text-to-image diffusion models, providing a scalable post-training framework to better align model generations with human preferences.
Within this Diffusion GRPO framework, scaling the rollout size has been shown to yield consistent and appreciable reward improvements~\cite{xue2025dancegrpo,expandgrpo}. By evaluating a massive candidate pool and selectively extracting only the most contrastive (e.g., the top and bottom) samples for model optimization, the GRPO objective function constructs a highly reliable gradient signal for stable and effective policy updates, leading to better alignment performance.

However, executing this rollout scaling paradigm on modern diffusion models~\cite{xie2024sana,flux2024,sd} imposes a substantial computation burden.
Because only a small set of highly contrastive samples is ultimately used for optimization, scaling the candidate pool shifts the training bottleneck from policy optimization to candidate generation.
To alleviate this bottleneck, we explore the integration of FP4 quantization into Diffusion RL rollout, effectively facilitating pronounced acceleration via quantized inference.
Yet, directly utilizing quantized rollout as optimization targets compromises the training performance and stability~\cite{flashrl_offpolicy,xi2026jetrlenablingonpolicyfp8}, imposing a restrictive ceiling on the performance of RL-based post-training.

To break this inherent dilemma between computational efficiency and training integrity, we propose \textbf{Sol-RL}, a \textbf{FP4-empowered Two-Stage Reinforcement Learning} framework. First, we reduce the cost of massive candidate generation by deploying FP4 quantization and reducing the number of sampling steps. Given the same initial noise seed, the accelerated approximations maintain intra-group ranking consistency with their high-precision counterparts. This critical property allows us to reliably filter the massive candidate pool and extract a high-variance subset that preserves the core relative advantage signals required by GRPO. Second, utilizing the highest- and lowest-ranked seeds identified in the first stage, we selectively regenerate this high-variance subset in BF16 precision. The policy model is then optimized strictly on these high-fidelity samples, avoiding the risks of performance degradation introduced by training with quantized rollouts. By structurally decoupling exploration from gradient optimization, our method effectively resolves the generation bottleneck while preserving the training fidelity effectively on par with the BF16 rollout pipeline.

The primary contributions of this work are summarized as follows:

\begin{itemize}
    \item \textbf{Characterizing the Rollout Scaling and its Bottleneck in Diffusion RL:} We demonstrate that scaling rollout candidate sizes and selective training on high-contrastive subsets yields pronounced alignment improvements, while shifting the primary training bottleneck from policy optimization to massive candidate generation.
    
    \item \textbf{Integration of FP4 in Diffusion RL Rollout:} We introduce FP4-empowered rollout into Diffusion Reinforcement Learning via a novel two-stage decoupled framework. By repurposing quantized rollout samples as an exploration proxy, we successfully scale rollout in Diffusion RL at a fraction of the computation cost.
    
    \item \textbf{Achieving Efficient Scaling without Sacrificing Alignment Quality:} Evaluated on diverse foundation models (SD3.5, FLUX.1, SANA) and reward metrics, our framework mitigates the efficiency-stability dilemma. It achieves up to $4.64\times$ convergence speedup while maintaining the alignment quality of the high-precision pipeline.
\end{itemize}

\section{Preliminaries}
\label{sec:preliminaries}

\paragraph{Group Relative Policy Optimization.}
In the simplest policy gradient formulation, REINFORCE~\citep{williams1992simple} optimizes
\begin{equation}
\nabla_\theta J(\theta)=\mathbb{E}_{\mathbf{x}\sim \pi_\theta(\cdot\mid c)}\!\left[\nabla_\theta \log \pi_\theta(\mathbf{x}\mid c)\, R(\mathbf{x},c)\right].
\end{equation}
Although unbiased, this Monte Carlo estimator typically exhibits high variance. A standard variance-reduction technique is to subtract a baseline $b(c)$~\citep{greensmith2004variance}, giving
\begin{equation}
\nabla_\theta J(\theta)=\mathbb{E}_{\mathbf{x}\sim \pi_\theta(\cdot\mid c)}\!\left[\nabla_\theta \log \pi_\theta(\mathbf{x}\mid c)\bigl(R(\mathbf{x},c)-b(c)\bigr)\right].
\end{equation}
Most modern algorithms, including PPO~\citep{schulman2017proximal}, implement this baseline using a learned value network (critic). Although effective, such critics introduce substantial memory overhead and may themselves become a source of training instability.

Group Relative Policy Optimization (GRPO)~\citep{shao2024deepseekmath} circumvents this by evaluating a group of candidate responses. For a given conditioning prompt $c$, the policy generates a group of $N$ independent rollout samples $\{\mathbf{x}^{(i)}\}_{i=1}^N$. Then it computes advantages using only the relative rewards within each sampled group. Specifically, given a reward function $R(\cdot)$, the advantage of the $i$-th sample is obtained by standardizing its reward against the group statistics:

\begin{equation}
A_i = \frac{R(\mathbf{x}^{(i)}) - \mu_R}{\sigma_R}, \quad \text{where} \quad \mu_R = \frac{1}{N}\sum_{j=1}^N R(\mathbf{x}^{(j)}), \quad \sigma_R = \sqrt{\frac{1}{N}\sum_{j=1}^N \left( R(\mathbf{x}^{(j)}) - \mu_R \right)^2}.
\end{equation}

Using these group-relative advantages, GRPO optimizes a PPO-style clipped surrogate objective over the sampled group, while regularizing the policy toward a reference model $\pi_{\text{ref}}$ through a direct KL term in the loss:

{\small
\begin{equation}
\mathcal{L}_{\text{GRPO}}(\theta) = \mathbb{E}_{\pi_{\text{old}}} \left[ \frac{1}{N} \sum_{i=1}^N \left( \min \left( r_i(\theta) A_i, \text{clip}\big(r_i(\theta), 1-\epsilon, 1+\epsilon\big) A_i \right) - \beta \mathbb{D}_{\text{KL}}\left(\pi_\theta(\cdot|c) \,\|\, \pi_{\text{ref}}(\cdot|c)\right) \right) \right]
\end{equation}
}

where $r_i(\theta) = \frac{\pi_\theta(\mathbf{x}^{(i)}|c)}{\pi_{\text{old}}(\mathbf{x}^{(i)}|c)}$ denotes the probability ratio of the policy distributions. 
This formulation makes the quality of the policy update strongly dependent on the sampled group, in particular on the informativeness of the candidate responses. Increasing $N$ can provide more informative within-group comparisons and more stable group statistics, but it also incurs substantially higher rollout and reward-evaluation costs during data collection.

\paragraph{FP4 Quantization and Hardware Acceleration.}
Driven by recent hardware advancements (e.g., NVIDIA Blackwell), 4-bit floating-point (FP4) arithmetic has emerged as a promising acceleration paradigm. FP4 encodes values using an extremely constrained bit-width (1 sign, 2 exponent, and 1 mantissa bit). To maintain numerical fidelity, it employs block-level micro-scaling, where contiguous elements share a single scaling factor. Specific implementations vary: the OCP MXFP4 standard groups 32 elements under an E8M0 scale, whereas NVIDIA's NVFP4 groups 16 elements under an E4M3 scale. Mathematically, the FP4 quantization of a high-precision tensor $\mathbf{x}$ is formulated as:$$\tilde{\mathbf{x}} = Q(\mathbf{x}) = S \cdot \Pi_{\text{FP4}}\left( \frac{\mathbf{x}}{S} \right),$$where $S$ is the shared scaling factor and $\Pi_{\text{FP4}}(\cdot)$ denotes the projection function. By leveraging these shared scales, FP4 achieves a massive $4\times$ throughput increase with minimal precision degradation.

\section{Methodology}
\label{sec:methodology}

\begin{figure}[t]
    \centering
    \includegraphics[width=\linewidth]{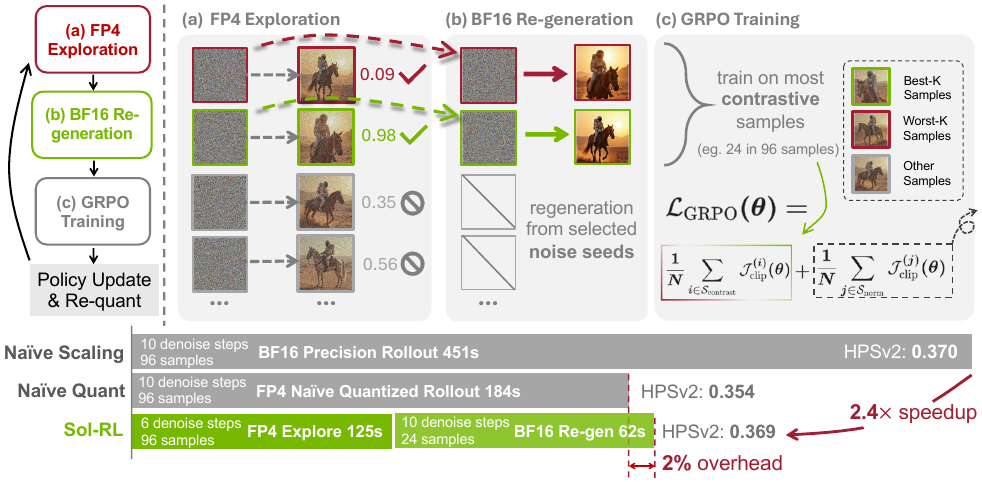}
    \caption{\textbf{Decoupled two-stage reinforcement learning pipeline of Sol-RL.} We separate the high-throughput FP4 exploration from the selective BF16 high-contrastive rollout. This framework achieves up to $2.4\times$ acceleration compared to naive scaling while avoiding quantization-induced corruption, introducing merely a 2\% computational overhead.}
    \label{fig:pipeline}
\end{figure}

Although expanding the pool of exploratory rollouts yields substantial performance gains, it creates a bottleneck in the Reinforcement Learning pipeline with heavy inference costs. The naive application of acceleration techniques often compromises the visual integrity of the generated samples and destabilizes the optimization process. To break this dilemma, we introduce a novel decoupled architecture that leverages FP4 quantization exclusively for high-throughput exploration while preserving high-fidelity rollout for policy optimization. 

In the remainder of this section, we first characterize the behavior of rollout scaling in Diffusion RL and identify its critical efficiency bottlenecks (Section~\ref{subsec:scaling}). Next, we investigate the FP4-quantized rollout samples, specifically analyzing the risks of employing them as direct optimization targets (Section~\ref{subsec:degradation}) alongside their potential utility as proxies for intra-group reward ranking estimation (Section~\ref{subsec:proxy}). Finally, we detail our decoupled FP4 exploration framework (Section~\ref{subsec:pipeline}), which achieves efficient and effective rollout scaling.

\subsection{Promise and Bottleneck of Rollout Scaling}
\label{subsec:scaling}

Recent advancements in reinforcement learning have compellingly demonstrated the immense value of scaling the number of rollouts per example~\citep{hu2025brorl}. Larger rollout groups broaden exploration scale and yield better samples, leading to better policy improvement. Beyond scaling rollout numbers, \citet{xue2025dancegrpo} propose a selective training framework that optimizes the policy using only the most contrastive samples from a massive rollout pool, e.g., the best k samples and the worst k samples. From a perspective of GRPO training dynamics, the most contrastive samples provide more reliable and informative learning signals for policy optimization, while other samples provide limited gradient due to the near-zero advantages. This paradigm scales rollout number while training overhead remains unchanged, bringing faster convergence and superior alignment performance.

However, the scaling of rollout shifts the computational bottleneck from policy optimization to candidate generation, as shown in Figure~\ref{fig:time_breakdown}. Moreover, under such selective training paradigm, only a highly contrastive fraction is selected for gradient updates and the remaining bulk of samples is discarded, revealing its inherent algorithmic redundancy. These compounding inefficiencies motivate the use of inference acceleration techniques, such as low-bit quantization.

\subsection{Training Degradation with Direct Quantized Rollouts}
\label{subsec:degradation}

As a widely used acceleration technique, quantization is a natural choice for reducing the computational cost of RL rollouts. However, recent studies demonstrate that directly utilizing these quantized rollout samples for RL optimization empirically leads to severe alignment degradation and training instabilities \cite{flashrl_offpolicy,xi2026jetrlenablingonpolicyfp8}. 
A key concern is the off-policy gap: trajectories sampled by quantized policy exhibit an inherent distribution shift from the high-precision target policy, potentially disrupting delicate policy updates.

Furthermore, for diffusion models post-training, the continuous nature of the state space exacerbates this degradation. Mainstream ``forward-process'' diffusion RL algorithms—especially Advantage Weighted Matching (AWM)~\citep{xue2025advantage}, as well as DiffusionNFT~\citep{diffusionnft}—formulate their objectives based on denoising score matching loss, treating the rollout samples as direct regression targets. As shown in Figure~\ref{fig:degradation}, when corrupted by low-bit quantization (e.g., FP4), the numerical noise forces the high-precision policy to mimic distorted, low-fidelity semantics. Consequently, this naive substitution inherently caps the achievable alignment quality of the model, finally neutralizing the benefits of rollout scaling.

\subsection{Proxy Reward Ranking via FP4 Exploration}
\label{subsec:proxy}

As demonstrated previously, because low-precision samples cannot faithfully match their high-precision counterparts at the pixel level, directly using them as training targets leads to training degradation. Consequently, we repurpose NVFP4 quantized rollout for a more error-tolerant objective in the reinforcement learning pipeline.

Our key observation is that, under the deterministic nature of ODE-style diffusion sampling, the coarse semantic layout and structural outcome of a generated rollout are fundamentally dictated by its initial noise, and this in turn influences the reward level of the sample~\cite{li2025mixgrpo, he2025tempflow, deng2026densegrposparsedensereward}. As visualized in Figure~\ref{fig:quantization_visual}, NVFP4 inference naturally preserves this semantic structure despite localized deviations, allowing us to use these high-throughput rollouts to rapidly estimate approximate reward magnitudes. This, in turn, enables us to reliably deduce the intra-group relative rankings among massive candidate seeds with minimal computational overhead.

\begin{figure}[htbp]
    \centering
    
    \begin{subfigure}[b]{0.298\linewidth}
        \centering
        \includegraphics[width=\linewidth]{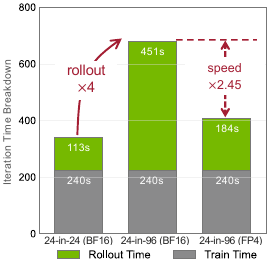} 
        \caption{Iteration Time Breakdown}
        \label{fig:time_breakdown}
    \end{subfigure}
    \hfill
    \begin{subfigure}[b]{0.30\linewidth}
        \centering
        \includegraphics[width=\linewidth]{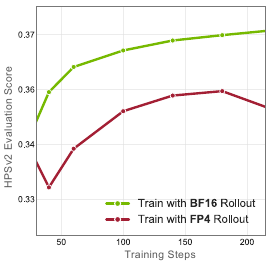} 
        \caption{Training Performance Comparison}
        \label{fig:degradation}
    \end{subfigure}
    \hfill
    \begin{subfigure}[b]{0.355\linewidth}
        \centering
        \includegraphics[width=\linewidth]{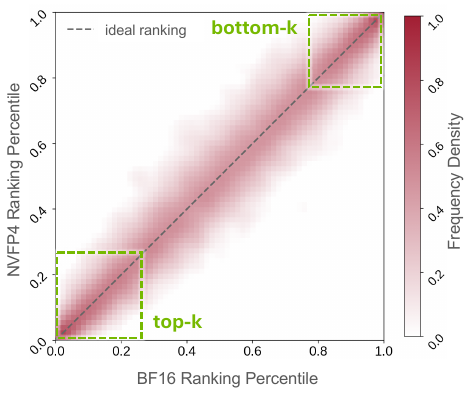}
        \caption{Proxy Ranking Reliability}
        \label{fig:correlation_c}
    \end{subfigure}
    
    \caption{\textbf{Pitfalls and Potential of NVFP4 rollouts.} (a) Time breakdown of high-precision rollout scaling and direct quantized rollout. The x-axis labels follow the format $K$-in-$N$ ($P$), denoting that $K$ samples are selected for training from $N$ generated rollouts under $P$ precision. (b) Directly integrating FP4 rollout in RL pipeline leads to severe instability and performance degradation compared to the BF16 baseline. (c) Conversely, the dense diagonal distribution of intra-group relative reward rankings validates NVFP4 quantized rollouts as a reliable proxy for reward sorting.}
    \label{fig:correlation}
\end{figure}

To empirically validate this, Figure~\ref{fig:correlation_c} presents a conditional probability density map comparing the true BF16 ranks against the NVFP4 proxy ranks.
Intuitively, given a sample's true rank percentile $x$, the vertical slice at $x$ reveals the probability distribution of its NVFP4 proxy rank.
The heavy concentration of probability density along the diagonal proves that NVFP4 rollouts accurately preserve the intra-group relative ordering, especially within the Top-K and Bottom-K quadrants, which are also the most contrastive samples essential for policy optimization.
In other words, NVFP4 rollouts serve as reliable and efficient proxies for intra-group reward rank. By utilizing these low-precision forward passes solely to evaluate and rank candidates, we can rapidly identify the specific seeds that are destined to yield highly contrastive rewards when subsequently regenerated in BF16 precision.

\subsection{FP4-Empowered Two-Stage Framework}
\label{subsec:pipeline}

Building upon the insight that FP4 rollouts serve as highly reliable proxies for relative reward ranking, we instantiate our decoupled design philosophy into \textbf{Sol-RL}, a novel \textbf{Two-Stage Rollout Pipeline}, as demonstrated in Figure~\ref{fig:pipeline}. This framework successfully harnesses the high throughput of NVFP4 quantized rollout while systematically circumventing the optimization degradation and instability inherent to training with quantized targets.

\paragraph{Stage 1: Accelerated Exploration at Scale via FP4.} 

As mentioned in Section~\ref{subsec:scaling}, while an expanded candidate pool dramatically improves exploration, our selective training paradigm only utilizes a small subset of highly contrastive samples as actual fitting targets. Consequently, computing the entire candidate pool in high precision (e.g., BF16) introduces algorithmic redundancy, as the vast majority of the generated samples are ultimately discarded. 

To reduce this massive computational waste, we construct the expanded pool by sampling $N$ independent initial noises $\{\mathbf{z}^{(i)}\}_{i=1}^N$ (e.g., $N=96$) and generating samples through a NVFP4 model ODE solver. To further maximize throughput, this proxy generation utilizes reduced inference steps (e.g., 6 steps) to rapidly compute their corresponding proxy rewards $\{\tilde{R}_i\}_{i=1}^N$. By deploying NVFP4 quantization, we unlock the extreme throughput potential of the latest hardware architectures, where NVFP4 dense operations deliver up to $4\times$ the TFLOPs of standard BF16 arithmetic. This highly efficient exploration provides a reliable proxy for relative reward ranking estimation, allowing us to accurately filter the scaled pool and isolate a minimal subset of $K$ high-contrastive seeds (i.e., the top and bottom candidates). 

\paragraph{Stage 2: High-Fidelity Regeneration and Policy Update.} 
The selected $K$ seeds (e.g., $K=24$) are then used to regenerate samples in the original high-precision (BF16) diffusion loop. By completely shielding the underlying vector field $v_\theta$ from low-precision quantization, this phase allows the ODE solver to reliably reconstruct high-fidelity samples $\mathbf{x}_0$ using the most contrastive noise seeds filtered during Stage 1. Subsequently, the policy network performs standard gradient-based optimization exclusively on these $K$ high-fidelity samples. By ensuring the generation process of training targets entirely in BF16, our pipeline substantially mitigates the risks of numerical instabilities typically associated with quantized rollout. Once the policy is updated, its weights are requantized into NVFP4 with negligible computational overhead and synchronized to the inference model for the next rollout iteration. 

In summary, this FP4-empowered decoupled exploration framework resolves the dilemma between training integrity and efficiency in diffusion RL. 
It harnesses the $4\times$ TFLOPs of FP4 to efficiently explore a massive candidate pool and identify its most contrastive samples, while reserving expensive high-precision compute strictly for the $K$ samples that actually dictate the policy update. Our approach successfully unlocks the superior alignment capabilities of scaled rollouts, while decoupling the overall computational bottleneck from the massive candidate size.

\section{Experiments}

\subsection{Experimental Setup}

We evaluate \textbf{Sol-RL}, our two-stage reinforcement learning framework, across three state-of-the-art text-to-image diffusion models: SANA~\cite{xie2024sana}, FLUX.1~\cite{flux2024}, and Stable Diffusion 3.5-Large (SD3.5-L)~\cite{sd}. We adopt the NVIDIA Transformer Engine~\cite{transformerengine} as our NVFP4 backend. All experiments are conducted on 8 NVIDIA B200 GPUs.

\textbf{Reward Models and Datasets.} We utilize ImageReward~\cite{xu2023imagereward}, CLIPScore~\cite{clipscore}, PickScore~\cite{kirstain2023pick}, and HPSv2~\cite{hpsv2} as our primary alignment objectives to measure visual quality and human preference. For the prompt dataset, we sample training and evaluation prompts from PickScore~\cite{kirstain2023pick}.

\textbf{Rollout Generation.} Our decoupled exploration relies on a highly efficient two-stage sampling mechanism. In the first stage, we generate an aggressively scaled exploration pool of 96 candidate samples per prompt using an NVFP4-compiled model in just 6 inference steps. We then isolate and preserve the initial noises of the most contrastive samples (specifically, the top-12 and bottom-12). In the second stage, we regenerate these 24 selected samples from the preserved initial noises using BF16 precision over 10 inference steps to construct high-fidelity rollouts.

\textbf{Training and Optimization.} The policy is optimized using the DiffusionNFT~\cite{diffusionnft} objective based on the 24 high-fidelity rollouts. The newly optimized weights are re-quantized and copied in-place into the compiled inference model after each update step, avoiding the computational overhead of recompilation during the iterative training loop. We apply Low-Rank Adaptation (LoRA)~\cite{hu2022lora} with a rank of $r=32$ and a scaling factor of $\alpha=64$ across all experiments.

\textbf{Baselines and Hyperparameters.} We compare our approach against reinforcement learning algorithms for diffusion models, including AWM~\cite{xue2025advantage}, DiffusionNFT~\cite{diffusionnft}, FlowGRPO~\cite{liu2025flow}, and DanceGRPO~\cite{xue2025dancegrpo}. To ensure a fair comparison, the majority of our hyperparameters are aligned with DiffusionNFT. Additional details are provided in the Appendix. 

\newcommand{\gdel}[1]{\textcolor{gray}{#1}}

\definecolor{bestbg}{HTML}{C8E6C9} 
\definecolor{secbg}{HTML}{E8F5E9}
\newcommand{\bestc}[1]{\cellcolor{bestbg}\textbf{#1}}
\newcommand{\secc}[1]{\cellcolor{secbg}#1}
\newcommand{\eqw}[1]{\makebox[1.5cm][c]{#1}}

\begin{table}[htbp]
\centering
\renewcommand{\arraystretch}{1.15}
\caption{\textbf{Quantitative comparison of alignment performance.} Evaluated on FLUX.1 under an identical GPU-hour budget. \gdel{$\Delta$} indicates the performance improvement over the Base (w/o CFG) model. \textbf{Bold} and darker green background indicate the best results, while lighter green background indicates the second best.}
\label{tab:main_results}
\resizebox{\linewidth}{!}{%
\begin{tabular}{l cc cc cc cc}
\toprule
\multirow{3}{*}{\textbf{Method}} & \multicolumn{2}{c}{\textbf{ImageReward}} & \multicolumn{2}{c}{\textbf{CLIPScore}} & \multicolumn{2}{c}{\textbf{PickScore}} & \multicolumn{2}{c}{\textbf{HPSv2}} \\
& \multicolumn{2}{c}{\gdel{\small (Base w/o CFG: 0.455)}} & \multicolumn{2}{c}{\gdel{\small (Base w/o CFG: 0.2630)}} & \multicolumn{2}{c}{\gdel{\small (Base w/o CFG: 0.8096)}} & \multicolumn{2}{c}{\gdel{\small (Base w/o CFG: 0.2566)}} \\
\cmidrule(lr){2-3} \cmidrule(lr){4-5} \cmidrule(lr){6-7} \cmidrule(lr){8-9}
& \eqw{Score} & \eqw{\gdel{$\Delta$ $(\uparrow)$}} & \eqw{Score} & \eqw{\gdel{$\Delta$ $(\uparrow)$}} & \eqw{Score} & \eqw{\gdel{$\Delta$ $(\uparrow)$}} & \eqw{Score} & \eqw{\gdel{$\Delta$ $(\uparrow)$}} \\
\midrule
DanceGRPO      & 1.4937 & \gdel{+1.0387} & 0.2898 & \gdel{+0.0268} & 0.8807 & \gdel{+0.0711} & 0.3552 & \gdel{+0.0986} \\
FlowGRPO       & 1.5331 & \gdel{+1.0781} & 0.2884 & \gdel{+0.0254} & 0.8743 & \gdel{+0.0647} & 0.3501 & \gdel{+0.0935} \\
AWM            & 1.6693 & \gdel{+1.2143} & \secc{0.3039} & \cellcolor{secbg}\gdel{+0.0409} & 0.8842 & \gdel{+0.0746} & \secc{0.3664} & \cellcolor{secbg}\gdel{+0.1098} \\
DiffusionNFT   & \secc{1.6707} & \cellcolor{secbg}\gdel{+1.2157} & 0.2991 & \gdel{+0.0361} & \secc{0.8852} & \cellcolor{secbg}\gdel{+0.0756} & 0.3613 & \gdel{+0.1047} \\
\textbf{Sol-RL (Ours)} & \bestc{1.7636} & \cellcolor{bestbg}\gdel{\textbf{+1.3086}} & \bestc{0.3089} & \cellcolor{bestbg}\gdel{\textbf{+0.0459}} & \bestc{0.8932} & \cellcolor{bestbg}\gdel{\textbf{+0.0836}} & \bestc{0.3688} & \cellcolor{bestbg}\gdel{\textbf{+0.1122}} \\
\bottomrule
\end{tabular}%
} 
\end{table}

\subsection{Main Results}

To substantiate these findings, we provide a detailed quantitative comparison on the FLUX.1~\cite{flux2024} base model in Table \ref{tab:main_results}. We benchmark our approach against FlowGRPO~\cite{liu2025flow}, DanceGRPO~\cite{xue2025dancegrpo}, AWM~\cite{xue2025advantage} and DiffusionNFT~\cite{diffusionnft}. As shown in the table, within the identical computational budget, our method consistently achieves superior alignment performance, demonstrating robust and comprehensive improvements across all evaluated metrics.

\begin{figure}[t]
\centering
\includegraphics[width=0.95\textwidth]{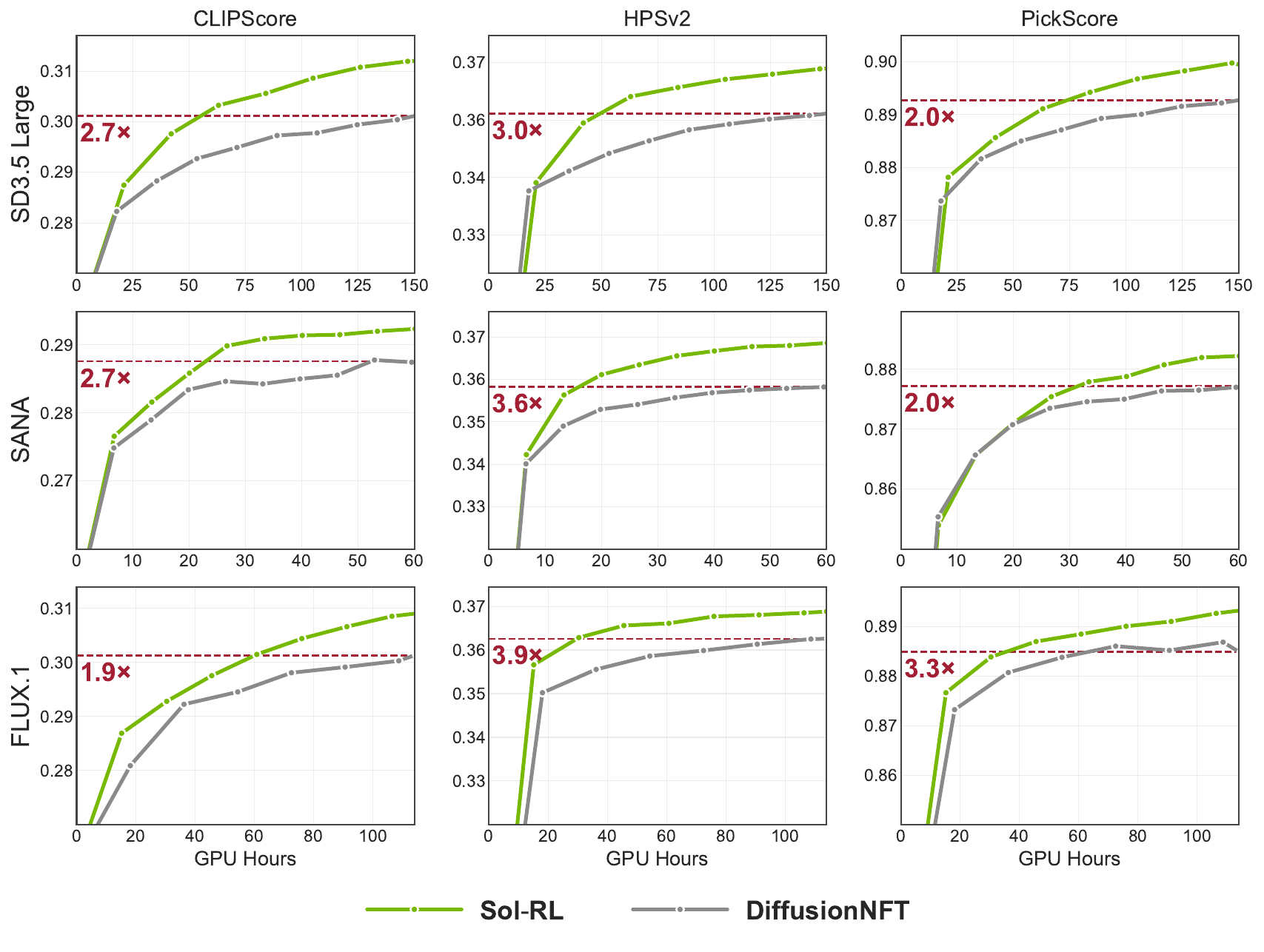}
\caption{\textbf{Comparison across diverse foundation models and alignment metrics.} Evaluated under identical wall-clock budgets (GPU Hours), \textbf{Sol-RL} (green) consistently outperforms the DiffusionNFT baseline (grey). Across all tested combinations of models and reward functions, our decoupled scaling strategy accelerates convergence to the baseline's equivalent performance by up to $4.64\times$, ultimately converging to a remarkably higher final alignment ceiling.}
\label{fig:main_results_visual}
\end{figure}

Figure~\ref{fig:main_results_visual} illustrates the overall alignment performance of the Sol-RL framework. Evaluated under identical GPU-hour budgets, our method consistently surpasses the DiffusionNFT~\cite{diffusionnft} baseline across diverse T2I foundation models and reward metrics. As demonstrated by the learning curves (Figure~\ref{fig:teaser} and~\ref{fig:main_results_visual}), Sol-RL accelerates convergence to the baseline's equivalent performance by $1.91\times$ to $4.64\times$, pushing the final alignment to a remarkably higher level.

\begin{figure}[t]
    \centering
    \includegraphics[width=\linewidth]{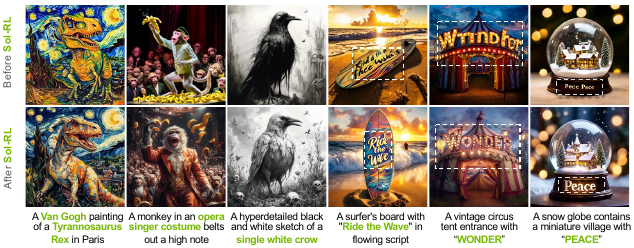}
\caption{\textbf{Visual comparison before and after Sol-RL.} Compared to the SANA base model without fine-tuning (top row), the counterpart optimized across multiple rewards (HPSv2, PickScore, CLIPScore and OCR) via Sol-RL (bottom row) exhibits substantial improvements in complex detail rendering and semantic alignment across various prompts. }
\label{fig:qualitative_results}
\end{figure}

To qualitatively evaluate the effectiveness of our approach, we present visual comparisons in Figure~\ref{fig:qualitative_results}. By optimizing the SANA across multiple rewards (ImageReward~\cite{xu2023imagereward}, CLIPScore~\cite{clipscore}, PickScore~\cite{kirstain2023pick}, HPSv2~\cite{hpsv2} and OCR~\cite{chen2023textdiffuser}), Sol-RL achieves substantial improvements in detail rendering and semantic alignment compared to the base model, demonstrating the effectiveness of our framework for comprehensive human preference alignment.

\subsection{Ablation Experiments}

To further unpack the alignment dynamics and proxy reliability within our framework, we ablate two critical hyperparameters: the number of FP4 exploration denoising steps ($T$) and the exploration pool size ($N$).

\textbf{Sensitivity of FP4 Exploration Denoising Steps.} During FP4 exploration, the number of denoising steps influences the reliability of the proxy reward ranking. As shown in Table~\ref{tab:ablation_steps}, employing an overly reduced number of steps (e.g., 4 steps) yields suboptimal alignment scores. This is because the coarse semantic layouts are insufficiently formed, leading to inaccurate intra-group ranking and suboptimal Top-K selection. Conversely, extending the exploration beyond $T=6$ shows no further improvement in the final reward, indicating that the proxy's ranking capability has already saturated. 

\textbf{Impact of Exploration Pool Size ($N$).} Scaling rollout can broaden the exploration space and enhance alignment quality~\cite{xue2025dancegrpo}. Serving as a highly efficient proxy for BF16 rollout, our decoupled exploration scaling replicates this favorable scaling behavior. As shown in Table~\ref{tab:ablation_poolsize}, increasing the exploration pool size $N \in \{24, 48, 72, 96\}$ while fixing the subset at $K=24$ yields continuous improvements in the final scores. This confirms that increasing the FP4 exploration pool size effectively unlocks substantial and consistent alignment gains.

\begin{table}[htbp]
    \centering
    \small
    \renewcommand{\arraystretch}{1.1}
    \begin{minipage}[t]{0.27\textwidth} 
        \centering
        \caption{\textbf{Increasing the exploration denoising steps $T$} progressively enhances the effectiveness of FP4 proxy reward sorting, leading to reward gains while saturating beyond 6 steps.}
        \label{tab:ablation_steps}
        \vspace{4pt}
        \setlength{\tabcolsep}{15pt}
        \begin{tabular}{cc}
            \toprule
            Steps ($T$) & \textbf{HPSv2} \\
            \midrule
            2 steps & 0.3587 \\
            4 steps & 0.3650 \\
            6 steps & \textbf{0.3686} \\
            8 steps & 0.3659 \\
            \bottomrule
        \end{tabular}
    \end{minipage}
    \hfill 
    \begin{minipage}[t]{0.27\textwidth} 
        \centering
        \caption{\textbf{Scaling the exploration pool size $N$} consistently improves alignment performance by providing a broader search space for discovering high-contrastive samples.}
        \label{tab:ablation_poolsize}
        \vspace{4pt}
        \setlength{\tabcolsep}{15pt}
        \begin{tabular}{cc}
            \toprule
            Size ($N$) & \textbf{HPSv2} \\
            \midrule
            $N=24$  & 0.3569 \\
            $N=48$  & 0.3622 \\
            $N=72$  & 0.3663 \\
            $N=96$  & \textbf{0.3686} \\
            \bottomrule
        \end{tabular}
    \end{minipage}
    \hfill 
    \begin{minipage}[t]{0.42\textwidth} 
        \centering
        \caption{\textbf{Quantitative evaluation of NVFP4 rollouts.} Compared with the uncompressed BF16 baseline, our accelerated NVFP4 rollouts achieve on-par Inception Score (IS) and CLIP scores across multiple base T2I models, proving that NVFP4 quantization maintains the semantic integrity.}
        \label{tab:quantization_fidelity}
        \vspace{4pt}
        \resizebox{\linewidth}{!}{
        \begin{tabular}{lcccc}
            \toprule
            \multirow{2}{*}{\textbf{Base Model}} & \multicolumn{2}{c}{\textbf{IS} ($\uparrow$)} & \multicolumn{2}{c}{\textbf{CLIP} ($\uparrow$)} \\
            \cmidrule(lr){2-3} \cmidrule(lr){4-5}
            & \textbf{BF16} & \textbf{NVFP4} & \textbf{BF16} & \textbf{NVFP4} \\
            \midrule
            FLUX.1      & 16.84 & 17.85 & 27.44 & 27.10 \\
            SANA        & 16.02 & 15.94 & 29.53 & 29.43 \\
            SD3.5-Large & 16.42 & 17.60 & 28.37 & 28.34 \\
            \bottomrule
        \end{tabular}
        }
    \end{minipage}

\end{table}

\vspace{-1em}
\subsection{Analysis of Two-Stage Decoupled Rollout}

\begin{wrapfigure}{r}{0.48\textwidth}
    \vspace{-12pt}
    \centering
    \includegraphics[width=\linewidth]{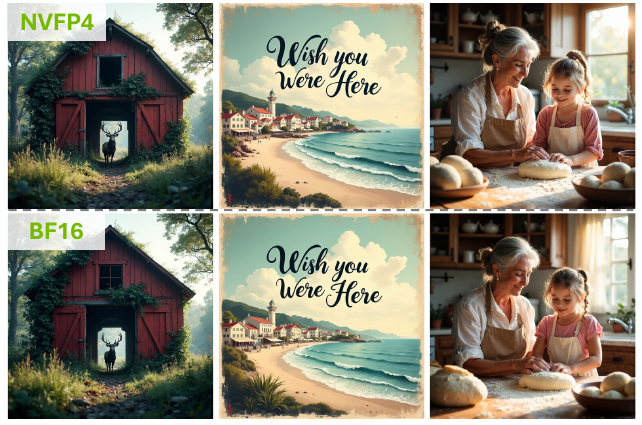} 
    \vspace{-10pt}
    \caption{\textbf{Visualization of NVFP4 and BF16 rollouts.} Despite minor localized deviations, the NVFP4 quantized rollouts maintain the overall semantic layout and structure.}
    \label{fig:quantization_visual}
    \vspace{-35pt}
\end{wrapfigure}

To further unpack the underlying mechanisms of our training framework, we provide a detailed breakdown of both the computational efficiency and the training integrity enabled by our decoupled architecture.

\paragraph{Analysis of NVFP4 Quantization Error.} As illustrated in Figure~\ref{fig:quantization_visual}, compared to BF16 baseline, NVFP4 quantization faithfully preserves the semantic structure despite localized deviations. To substantiate these visual observations, Table~\ref{tab:quantization_fidelity} provides supporting quantitative evidence, including Inception Score and CLIPScore. It demonstrates that the semantics are robustly maintained under low-bit compression. These findings confirm that quantized rollout firmly preserves the necessary structural integrity compared to BF16 rollout, fully validating its reliability as a proxy for accurate intra-group reward ranking.

\paragraph{Training Efficiency Breakdown.} 
To precisely isolate the source of our acceleration, we evaluate the computational cost under the standard 24-in-96 rollout setting. We distinguish between the isolated \textit{Rollout Time} (the pure forward generation phase) and the overall \textit{Iteration Time} (the complete RL step, including both rollout and the subsequent backward gradient updates). As detailed in Table~\ref{tab:efficiency}, expanding the candidate pool to 96 under a BF16 rollout pipeline imposes a severe computational bottleneck. By accelerating the massive exploration phase via NVFP4 quantized rollout, Sol-RL achieves up to a $2.4\times$ speedup in the isolated rollout phase and a $1.6s\times$ acceleration in the overall end-to-end iteration time, confirming that our framework successfully unlocks massive rollout scaling. 

\begin{table}[htbp]
    \centering
    \small
    \renewcommand{\arraystretch}{1.15}
    \begin{minipage}[t]{0.58\textwidth}
        \centering
        \caption{\textbf{Training efficiency and acceleration analysis.} We compare the exact time consumption of the naive scaling rollout with our Two-stage \textbf{Sol-RL} framework. Our method reduces the generation overhead, achieving up to a \textbf{2.41$\times$} speedup in rollout time and a \textbf{1.62$\times$} acceleration in overall end-to-end training, substantially alleviating the computational bottleneck of large-scale exploration.}
        \label{tab:efficiency}
        
        \vspace{4pt}
        \resizebox{\linewidth}{!}{ 
        \begin{tabular}{lcccccc}
        \toprule
        \multirow{2}{*}{Base Model} & \multicolumn{3}{c}{Rollout Time (s)} & \multicolumn{3}{c}{End-to-End Time (s)} \\
        \cmidrule(lr){2-4} \cmidrule(lr){5-7}
        & Naive & Ours & Speedup & Naive & Ours & Speedup \\
        \midrule
        FLUX.1      & 184 & 79  & \textbf{2.33$\times$} & 274 & 169 & \textbf{1.62$\times$} \\
        SD3.5-Large & 451 & 187 & \textbf{2.41$\times$} & 691 & 427 & \textbf{1.61$\times$} \\
        SANA        & 65  & 46  & \textbf{1.41$\times$} & 95  & 76  & \textbf{1.25$\times$} \\
        \bottomrule
        \end{tabular}
        }
    \end{minipage}\hfill
    \begin{minipage}[t]{0.39\textwidth}
        \centering
        \caption{\textbf{Alignment performance preservation.} Post-RL performance compared to the BF16 naive scaling rollout on HPSv2. Under identical training steps, our method maintains on-par results with a marginal gap ($\Delta$) of at most 1\% while achieving higher efficiency.}
        \vspace{-3pt}
        \label{tab:preservation}
        
        \vspace{5.5pt}
        \resizebox{0.97\linewidth}{!}{ 
        \begin{tabular}{lcc}
        \toprule
        \multirow{2}{*}{Base Model} & \multicolumn{2}{c}{HPSv2} \\
        \cmidrule(lr){2-3}
        & Naive & Sol-RL ($\Delta$) \\
        \midrule
        FLUX.1      & 0.3699 & 0.3688 (\gdel{$-0.29\%$}) \\
        SD3.5-Large & 0.3803 & 0.3762 (\gdel{$-1.08\%$}) \\
        SANA        & 0.3682 & 0.3686 (\gdel{$+0.11\%$}) \\
        \bottomrule
        \end{tabular}
        }
    \end{minipage}
    
\end{table}

\paragraph{Preservation of Alignment Fidelity.} 
In Sol-RL, we separate the policy optimization phase from the quantized exploration. As shown in Table~\ref{tab:preservation}, by updating the policy exclusively on the BF16 re-generated subset, our framework maintains the alignment fidelity of the naive scaling baseline (using BF16 precision brute-force sampling 96 images), while achieving remarkable acceleration as shown in Table~\ref{tab:efficiency}. This decoupled approach ensures stable policy optimization, effectively achieving high-throughput exploration efficiency without compromising the final generation quality.

\section{Related Work}

\subsection{Reinforcement Learning for Diffusion Models}

ImageReward~\citep{xu2023imagereward} introduces Reward Feedback Learning (ReFL), which directly maximizes the reward of an approximately one-step predicted image. To reduce the approximation error of one-step prediction, DRaFT~\citep{clark2023directly} instead optimizes rewards on final multi-step sampled images, and mitigates the resulting memory overhead through truncated backpropagation and gradient checkpointing. From a continuous-time perspective, Adjoint Matching~\citep{domingo2024adjoint} applies the adjoint method~\citep{pontryagin2018mathematical} for memory-efficient gradient computation. Reward Feedback Learning has also been explored on few-step distilled models~\citep{kim2024pagoda,li2024reward,luo2024diff,luo2025reward}.

DDPO~\citep{black2023training} and DPOK~\citep{fan2023dpok} formulate diffusion RL as a multi-step decision-making problem based on an Euler-Maruyama discretization of the reverse process, which yields a tractable Gaussian likelihood at each step. DeepSeekMath~\citep{shao2024deepseekmath} introduces Group Relative Policy Optimization (GRPO), which replaces the value-model baseline in PPO~\citep{schulman2017proximal} with the group-wise mean reward. Building on this line, Flow-GRPO~\citep{liu2025flow} and DanceGRPO~\citep{xue2025dancegrpo} combine the DDPO formulation with GRPO for diffusion model post-training. Recent studies have further improved sampling efficiency through trajectory branching~\citep{he2025tempflow}, mixed ODE/SDE sampling~\citep{li2025mixgrpo}, and structured denoising strategies~\citep{li2025branchgrpo,treegrpo,expandgrpo,fu2025dynamictreerpobreakingindependenttrajectory}.

Another line of work directly optimizes the forward diffusion process. \citet{lee2023aligning} fine-tune text-to-image models by maximizing an offline reward-weighted denoising loss, while \citet{fan2025online} extend this objective to an online setting with Wasserstein-2 regularization. Relatedly, Diffusion-DPO~\citep{wallace2024diffusion} offers a preference-optimization counterpart to this line, adapting DPO-style learning to diffusion model post-training without explicit rollouts. FMPG~\citep{mcallister2025flow} and especially AWM~\citep{xue2025advantage} place this line on firmer policy-optimization footing by using the ELBO as a proxy for policy likelihood. This connection makes forward-process optimization a particularly compelling direction. DiffusionNFT~\citep{diffusionnft} can be interpreted as an NFT-style~\citep{chen2025bridging} forward-process version of GRPO. Other works also explore forward-process variants for diffusion post-training~\citep{chen2025towards,luo2025reinforcing}, while~\citet{choi2026rethinking} provide a discussion of forward-based diffusion RL.

\subsection{Efficient Inference with Low-bit Quantization}

Model quantization has become a mainstream technique for deploying large foundation models. For Large Language Models (LLMs), early breakthroughs primarily focused on 8-bit integer (INT8) quantization~\cite{llmint8,smoothquant}. For 4-bit quantization, methods such as GPTQ and AWQ~\cite{gptq,awq}, leverage second-order Hessian information and activation-aware scaling to maintain high fidelity. Several studies push the boundary to 2-3 bits via techniques like learnable equivalent transformations and sparse-quantized representations~\cite{omniquant,aqlm,spqr}. Among these, recent work~\cite{llmfp4} utilizes the exact FP4 format implemented in the NVIDIA Blackwell architecture, achieving remarkable precision without introducing complex mechanism designs. Quantization of diffusion models is also extensively explored for inference acceleration. To address distribution shifts across denoising timesteps, early works~\cite{posttrainingquantizationdiffusionmodels,qdiffusion,ptqd} designed timestep-aware calibration and correlation-based noise correction. Recent advancements like SVDQuant~\cite{svdquant} have successfully bridged the gap to 4-bit inference by absorbing activation outliers through Singular Value Decomposition (SVD). 

Quantized inference has been introduced into reinforcement learning to alleviate the massive computational bottleneck. Frameworks such as FlashRL and QeRL~\cite{flashrl,qerl} have demonstrated substantial speedups via quantized rollout. Concurrent studies~\cite{flashrl_offpolicy,xi2026jetrlenablingonpolicyfp8} highlight that utilizing quantized inference for sampling shifts the optimization process into an off-policy setting, which may induce severe numerical discrepancies. To mitigate these off-policy vulnerabilities, QuRL~\cite{qurl} proposes adaptive clipping mechanisms to prevent divergence between the quantized actor and the BF16 precision policy. FP8-RL~\cite{fp8-rl} utilizes importance ratio between quantized inference and BF16 precision training. VESPO~\cite{vespo} introduces soft policy optimization to stabilize off-policy learning under such engine mismatches. Alternatively, at the system level, frameworks like Jet-RL~\cite{xi2026jetrlenablingonpolicyfp8} advocate for a unified FP8 precision flow across both the training and rollout phases, thereby fundamentally eliminating the off-policy gap and ensuring robust convergence.

\section{Conclusion}
\label{sec:conclusion}

In this work, we identified a critical efficiency-stability dilemma in diffusion reinforcement learning: while extensive rollout scaling serves as an effective mechanism for deriving more reliable and robust gradient signals, the immense computational cost of generation bottlenecks the training pipeline. To accelerate this process, we introduce NVFP4 quantization for efficient rollout; yet, we observed that directly utilizing these low-bit quantized samples for policy optimization potentially leads to alignment degradation and optimization instability. To address this challenge, we proposed a novel two-stage decoupled rollout framework. By strictly confining the high-throughput NVFP4 generation to an initial large-scale exploration phase, and reserving BF16 compute exclusively for regenerating the selected high-contrastive samples, our framework successfully decouples exploration efficiency from optimization stability. By seamlessly integrating the algorithmic mechanisms of rollout scaling and selective training with the system-level throughput gains of NVFP4, this approach creates a powerful synergy between the training strategy and hardware acceleration. Consequently, our approach achieves substantial acceleration up to 4.64$\times$ while maintaining robust alignment quality, effectively matching the training fidelity and generative performance of standard higher-precision pipeline without the extensive computational burden.

\newpage
\appendix
\onecolumn

\section{Theoretical Justification}
\label{app:theory}

The efficacy of FP4 exploration hinges on whether low-precision rollout can accurately preserve the relative reward ranking of candidates. We establish this rigorous guarantee by analyzing the worst-case perturbation bounds of the ODE solver through the lens of Extreme Value Theory (EVT).

\paragraph{Low Precision as a Bounded Perturbation.}
Let the high-precision trajectory satisfy the exact vector field $\dot{\mathbf{x}}_t = v_\theta(\mathbf{x}_t, t)$, and let the low-precision accelerated trajectory satisfy:
\begin{equation}
\dot{\tilde{\mathbf{x}}}_t = v_\theta(\tilde{\mathbf{x}}_t, t) + \mathbf{e}_t,
\end{equation}
where $\mathbf{e}_t$ denotes the effective perturbation induced by FP4 rounding errors and low-precision solver arithmetic. 

Assuming the vector field $v_\theta(\cdot, t)$ is $L_v$-Lipschitz continuous with respect to $\mathbf{x}$, we can apply standard comparison arguments and Gr\"onwall's inequality to bound the final sample deviation:
\begin{equation}
\|\mathbf{x}_0 - \tilde{\mathbf{x}}_0\| \le e^{L_v T}\int_0^T \|\mathbf{e}_s\|\, ds.
\end{equation}
Furthermore, assuming the alignment reward model $R(\mathbf{x})$ is $L_R$-Lipschitz, the absolute reward error for a fixed seed is strictly bounded by the inequality described below:
\begin{equation}
|R(\mathbf{x}_0) - R(\tilde{\mathbf{x}}_0)| \le L_R \|\mathbf{x}_0 - \tilde{\mathbf{x}}_0\| \le L_R e^{L_v T}\int_0^T \|\mathbf{e}_s\|\, ds =: \Delta.
\end{equation}
The quantity $\Delta$ establishes a theoretical upper bound on the cross-precision reward discrepancy for any given initial noise. Crucially, $\Delta$ is a static constant determined purely by the numerical precision format and the integration steps, independent of the candidate pool size $N$.

\paragraph{Extreme Value Guarantee under Rollout Scaling.}
Rollout-scaling Group-relative RL algorithms (e.g., GRPO) rely fundamentally on the strength of contrastive learning signal. Therefore, we evaluate the theoretical \textit{range} (the difference between the maximum and minimum rewards) preserved by our low-precision exploration.

For a given prompt, we model the true oracle rewards of the generated candidates as identically distributed from a sub-Gaussian distribution, $R \sim \mathcal{N}(\mu, \sigma^2)$. Let $R^*_{max} = \max_{i} R_i$ and $R^*_{min} = \min_{i} R_i$ denote the true maximum and minimum rewards within a scaled pool of size $N$. Using the classical extreme-value asymptotics for i.i.d. Gaussian samples, the expected true range $W^*_N = R^*_{max} - R^*_{min}$ expands symmetrically with $N$:
\begin{equation}
\mathbb{E}[W^*_N] = \mathbb{E}[R^*_{max}] - \mathbb{E}[R^*_{min}] \approx 2\sigma \sqrt{2 \log N}.
\end{equation}

Now, consider our accelerated exploration pipeline. The system observes the proxy rewards $\tilde{R}_i = R_i + \epsilon_i$, where the static quantization disturbance is tightly bounded by $|\epsilon_i| \le \Delta$. The screening mechanism selects the empirical best candidate $\hat{i}_{max} = \arg\max_i \tilde{R}_i$ and the empirical worst candidate $\hat{i}_{min} = \arg\min_i \tilde{R}_i$.  Combining these bounds yields the true reward range $\hat{W}$ of the empirically selected candidates:
\begin{equation}
\hat{W} = R_{\hat{i}_{max}} - R_{\hat{i}_{min}} \ge (R^*_{max} - 2\Delta) - (R^*_{min} + 2\Delta) = W^*_N - 4\Delta.
\end{equation}

Taking the expectation of this lower bound explicitly connects the retained gradient signal to the rollout scale $N$:
\begin{equation}
\mathbb{E}[\hat{W}] \ge 2\sigma \sqrt{2 \log N} - 4\Delta.
\end{equation}
The worst-case penalty incurred by low-precision exploration is a static constant contraction ($-4\Delta$) on the reward margin. However, the true extreme value advantage ($2\sigma \sqrt{2 \log N}$) grows monotonically with the scale $N$. As we aggressively scale up the rollout group, the extreme contrastive bounds of the distribution inevitably overpower the constant quantization noise, preserving the critical gradient signals required to unlock oracle alignment.

\section{Implementation Details}
\label{app:implementation}

\subsection{Training Hyperparameters}

Table~\ref{tab:hyperparams} summarizes the training hyperparameters for all three diffusion models. All rollouts use deterministic ODE sampling; classifier-free guidance is disabled for SANA and SD3.5, while FLUX.1 passes a guidance embedding of~$1.0$ to its transformer. Rows with a single value in the rightmost column are shared across all three models.

\begin{table}[t]
\centering
\caption{\textbf{Training hyperparameters.} Rows with a single value in the rightmost column are shared across all models.}
\label{tab:hyperparams}
\resizebox{\textwidth}{!}{%
\begin{tabular}{lcccc}
\toprule
\textbf{Category} & \textbf{Hyperparameter} & \textbf{SANA-1.5 1600M} & \textbf{FLUX.1-dev} & \textbf{SD 3.5 Large} \\
\midrule
\multirow{3}{*}{Model}
 & Image resolution        & $1024\!\times\!1024$ & $512\!\times\!512$ & $1024\!\times\!1024$ \\
 & Gradient checkpointing  & \xmarker & \cmarker & \xmarker \\
 & LoRA target modules     & \texttt{to\_\{q,k,v,out\}} & \texttt{to\_\{q,k,v,out\}} & \texttt{attn.\{to,add\}\_\{q,k,v,out\}} \\
\midrule
\multirow{3}{*}{LoRA}
 & Rank $r$                & \multicolumn{3}{c}{32} \\
 & Alpha $\alpha$          & \multicolumn{3}{c}{64} \\
 & Init mode               & \multicolumn{3}{c}{Gaussian} \\
\midrule
\multirow{6}{*}{Optimizer}
 & Algorithm               & \multicolumn{3}{c}{AdamW} \\
 & Learning rate           & \multicolumn{3}{c}{$3\!\times\!10^{-4}$} \\
 & $(\beta_1,\,\beta_2)$   & \multicolumn{3}{c}{$(0.9,\;0.999)$} \\
 & Weight decay            & \multicolumn{3}{c}{$1\!\times\!10^{-4}$} \\
 & $\epsilon$              & \multicolumn{3}{c}{$1\!\times\!10^{-8}$} \\
 & Mixed precision         & \multicolumn{3}{c}{BF16} \\
\midrule
\multirow{3}{*}{Rollout}
 & ODE solver              & Euler (flow) & DPM-Solver-2 & DPM-Solver-2 \\
 & Rollout steps           & 10 & 10 & 10 \\
 & Eval steps              & 40 & 28 & 40 \\
\midrule
\multirow{5}{*}{Training}
 & Per-GPU micro-batch     & 16 & 12 & 4 \\
 & Grad.\ accum.\ steps   & 9 & 12 & 36 \\
 & Timestep fraction       & 0.6 & 0.4 & 0.6 \\
 & Num.\ train timesteps   & 6 & 4 & 6 \\
 & Max gradient norm       & 1.0 & 1.0 & 0.002 \\
\midrule
\multirow{3}{*}{Loss}
 & guidance parameter $\beta$    & \multicolumn{3}{c}{$1.0$} \\
 & KL penalty $\beta_\mathrm{kl}$ & \multicolumn{3}{c}{$1\!\times\!10^{-4}$} \\
 & Advantage clip          & \multicolumn{3}{c}{$5$} \\
\midrule
\multirow{4}{*}{Best-of-$N$}
 & Prompts per epoch       & \multicolumn{3}{c}{48} \\
 & GPUs                    & \multicolumn{3}{c}{8} \\
 & Best-of-$N$ ($K$)       & \multicolumn{3}{c}{24} \\
 & Images per prompt ($N$) & \multicolumn{3}{c}{96} \\
\midrule
\multirow{3}{*}{\shortstack[l]{Two-stage\\Rollout}}
 & Exploration steps             & \multicolumn{3}{c}{6} \\
 & Exploration model             & \multicolumn{3}{c}{Compiled + NVFP4} \\
 & Full rollout model      & \multicolumn{3}{c}{Compiled (BF16)} \\
\midrule
\multirow{2}{*}{Regularization}
 & EMA decay               & \multicolumn{3}{c}{0.9} \\
 & Old-model decay         & \multicolumn{3}{c}{Linear ramp (rate $0.001$, cap $0.5$)} \\
\bottomrule
\end{tabular}%
}
\end{table}

\subsection{Rollout Pipeline Details}

The Sol-RL two-stage pipeline operates as follows during each training iteration. In Stage~1 (FP4 Exploration), the policy weights are first quantized into NVFP4 via the NVIDIA Transformer Engine and deployed onto a pre-compiled inference engine. For each prompt in the batch, $N = 96$ independent initial noise vectors are drawn, and the NVFP4 model generates candidate images with a reduced number of denoising steps ($T = 6$). Each candidate is scored by the reward model, and the top-$K/2$ and bottom-$K/2$ noise seeds are retained based on their proxy reward rankings. In Stage~2 (BF16 Regeneration), these $K = 24$ selected seeds are fed into the BF16 policy model with the full inference step budget ($T = 10$) to produce high-fidelity samples. The policy is then updated using the DiffusionNFT objective on this contrastive subset. After the gradient update, the new policy weights are re-quantized in-place into the NVFP4 inference engine without recompilation, preparing for the next iteration.

\subsection{Reward Models and Evaluation}
\label{app:reward}

We employ four widely used reward models as alignment objectives:
\begin{itemize}[leftmargin=1.5em, itemsep=2pt]
    \item \textbf{ImageReward}~\citep{xu2023imagereward}: A BLIP-based model trained on human preference annotations for overall visual quality.
    \item \textbf{CLIPScore}~\citep{clipscore}: The cosine similarity between CLIP text and image embeddings for evaluating semantic alignment.
    \item \textbf{PickScore}~\citep{kirstain2023pick}: A preference model trained on the Pick-a-Pic dataset, reflecting preference between image pairs.
    \item \textbf{HPSv2}~\citep{hpsv2}: Human Preference Score v2, a fine-tuned CLIP-based scorer trained on large-scale preference data.
\end{itemize}
For the prompt dataset, we sample prompts from the PickScore~\citep{kirstain2023pick} training split for RL training and hold out a separate subset for evaluation. During training, each reward model is used independently as the alignment objective; evaluation is performed on the held-out set using all four metrics.

\section{Additional Analysis of NVFP4 Exploration}
\label{app:analysis}

A core assumption of our decoupled framework is that while NVFP4 quantization may introduce slight perturbations to absolute reward values, it faithfully preserves the intra-group relative rankings of the generated candidates.

First, we evaluate the global ranking consistency using two standard non-parametric metrics: Kendall's $\tau$ and Spearman's $\rho$. Spearman's $\rho$ evaluates how well the relationship between two ranked variables can be described using a monotonic function, whereas Kendall's $\tau$ measures the ordinal association based on the ratio of concordant to discordant pairs. A Spearman's $\rho$ exceeding $0.80$ is widely regarded as indicating a very strong positive correlation, and a Kendall's $\tau$ above $0.70$ reflects highly consistent pairwise orderings. As reported in Table~\ref{tab:ranking_consistency}, our NVFP4 proxy rewards consistently exceed these rigorous thresholds, achieving an impressive average $\rho$ of $0.927$ and $\tau$ of $0.798$ across all reward models. This substantiates that the overall candidate distribution remains structurally intact under FP4 compression.

Furthermore, beyond global correlations, the efficiency of selective training heavily relies on the accurate identification of extreme samples---the best and worst candidates that provide the most substantial positive and negative advantage signals. To this end, we introduce the Top/Bottom-$k$ Match metric, which calculates the exact intersection rate of the highest and lowest $k$ items selected under BF16 versus NVFP4. A high Top-$k$ match indicates that the optimal candidates are successfully retained, while a low Bottom-$k$ false-inclusion rate ensures that poor candidates are not erroneously selected for optimization. Our results confirm that the NVFP4 proxy behaves as an exceptional filter: it accurately captures the most critical contrastive candidates with an over $96\%$ Top-4 precision and less than $4\%$ Bottom-4 false inclusion. These comprehensive metrics theoretically and empirically justify our design choice to aggressively scale exploration in NVFP4 while reserving BF16 exclusively for the optimization phase.

\begin{table}[htbp]
\centering
\caption{\textbf{Group-Relative Ranking Consistency.} Global reward correlation ($\tau$, $\rho$) and exact match rates (Top/Bottom $k$) between FP4-accelerated samples and high-fidelity BF16 baselines. These results demonstrate that our acceleration preserves intra-group relative rankings, thereby ensuring the effectiveness of FP4-driven exploration.}
\label{tab:ranking_consistency}
\resizebox{\textwidth}{!}{
\begin{tabular}{lcccccccc}
\toprule
\multirow{2}{*}{\textbf{Reward Metric}} & \multirow{2}{*}{\textbf{Kendall $\tau$ ($\uparrow$)}} & \multirow{2}{*}{\textbf{Spearman $\rho$ ($\uparrow$)}} & \multicolumn{2}{c}{\textbf{Top/Btm 4 Match}} & \multicolumn{2}{c}{\textbf{Top/Btm 8 Match}} & \multicolumn{2}{c}{\textbf{Top/Btm 12 Match}} \\
\cmidrule(lr){4-5} \cmidrule(lr){6-7} \cmidrule(lr){8-9}
& & & \textbf{Top ($\uparrow$)} & \textbf{Btm ($\downarrow$)} & \textbf{Top ($\uparrow$)} & \textbf{Btm ($\downarrow$)} & \textbf{Top ($\uparrow$)} & \textbf{Btm ($\downarrow$)} \\
\midrule
CLIPScore   & 0.752 & 0.900 & 95.7\% & 4.5\% & 93.9\% & 6.2\% & 92.2\% & 8.2\% \\
HPSv2       & 0.827 & 0.943 & 97.6\% & 3.4\% & 95.5\% & 5.3\% & 93.9\% & 7.1\% \\
ImageReward & 0.807 & 0.932 & 97.2\% & 3.9\% & 95.1\% & 5.9\% & 93.4\% & 7.6\% \\
PickScore   & 0.806 & 0.934 & 97.1\% & 3.8\% & 95.4\% & 5.6\% & 93.6\% & 7.2\% \\
\midrule
\textbf{Overall} & \textbf{0.798} & \textbf{0.927} & \textbf{96.9\%} & \textbf{3.9\%} & \textbf{95.0\%} & \textbf{5.7\%} & \textbf{93.3\%} & \textbf{7.5\%} \\
\bottomrule
\end{tabular}
}
\end{table}

\begin{figure}[t]
\centering
\includegraphics[width=0.95\textwidth]{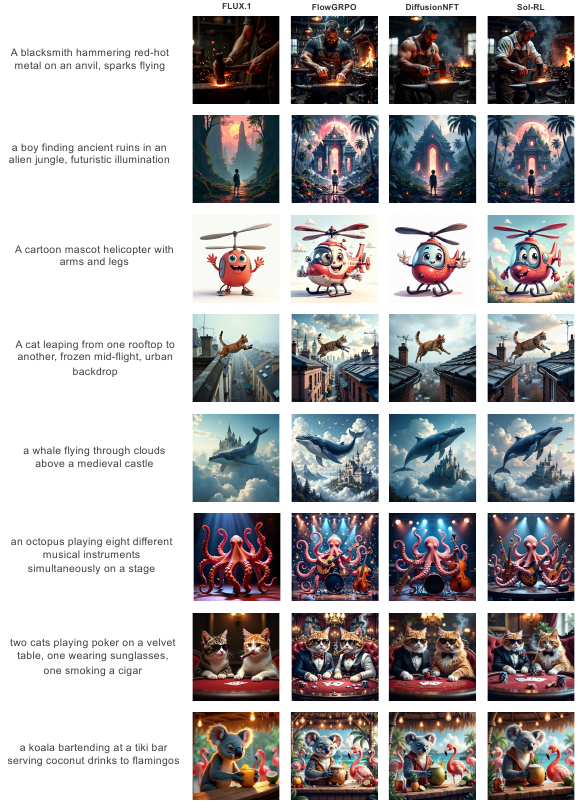}
\caption{\textbf{Qualitative comparison on PickScore-optimized models.} We compare images generated by Flux.1-dev base models against their Sol-RL, DiffusionNFT and FlowGRPO fine-tuned variants. Sol-RL produces images with stronger semantic alignment to the prompt, richer fine-grained details, and more coherent artistic style.}
\label{fig:demo_pickscore}
\end{figure}

\begin{figure}[t]
\centering
\includegraphics[width=0.97\textwidth]{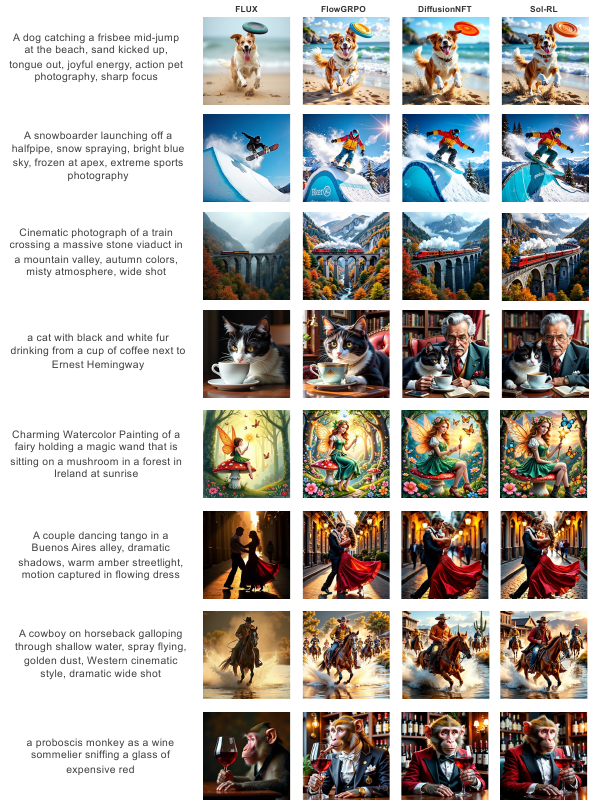}
\caption{\textbf{Qualitative comparison on HPSv2-optimized models.} We compare images generated by Flux.1-dev base models against their Sol-RL, DiffusionNFT and FlowGRPO fine-tuned variants. Sol-RL produces images with stronger semantic alignment to the prompt, richer fine-grained details, and more coherent artistic style.}
\label{fig:demo_hpsv2}
\end{figure}

\begin{figure}[t]
\centering
\includegraphics[width=1\textwidth]{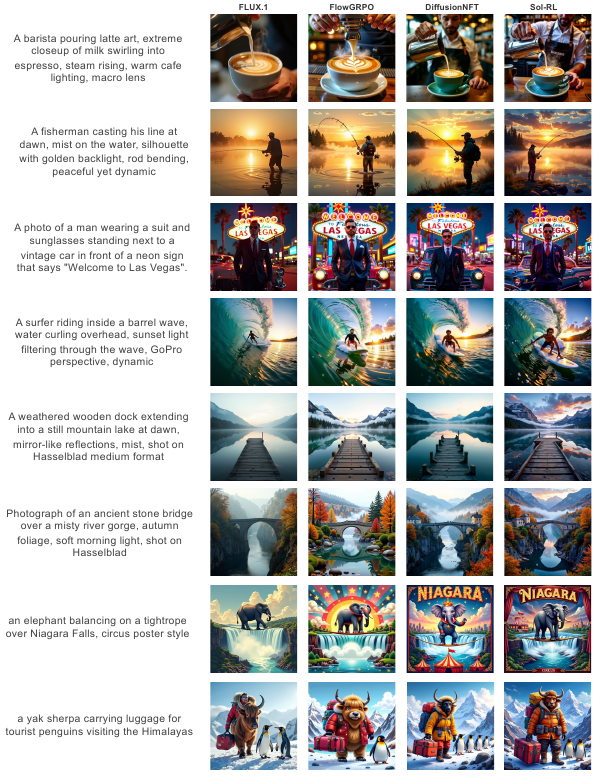}
\caption{\textbf{Qualitative comparison on ImageReward-optimized models.} We compare images generated by Flux.1-dev base models against their Sol-RL, DiffusionNFT and FlowGRPO fine-tuned variants. Sol-RL produces images with stronger semantic alignment to the prompt, richer fine-grained details, and more coherent artistic style.}
\label{fig:demo_imagereward}
\end{figure}

\clearpage

{
  \small
  \bibliographystyle{unsrtnat}
  \bibliography{ref}

@article{xue2025advantage,
  title={Advantage weighted matching: Aligning rl with pretraining in diffusion models},
  author={Xue, Shuchen and Ge, Chongjian and Zhang, Shilong and Li, Yichen and Ma, Zhi-Ming},
  journal={arXiv preprint arXiv:2509.25050},
  year={2025}
}

@misc{diffusionnft,
      title={DiffusionNFT: Online Diffusion Reinforcement with Forward Process}, 
      author={Kaiwen Zheng and Huayu Chen and Haotian Ye and Haoxiang Wang and Qinsheng Zhang and Kai Jiang and Hang Su and Stefano Ermon and Jun Zhu and Ming-Yu Liu},
      year={2026},
      eprint={2509.16117},
      archivePrefix={arXiv},
      primaryClass={cs.LG},
      url={https://arxiv.org/abs/2509.16117}, 
}

@misc{llmint8,
      title={LLM.int8(): 8-bit Matrix Multiplication for Transformers at Scale}, 
      author={Tim Dettmers and Mike Lewis and Younes Belkada and Luke Zettlemoyer},
      year={2022},
      eprint={2208.07339},
      archivePrefix={arXiv},
      primaryClass={cs.LG},
      url={https://arxiv.org/abs/2208.07339}, 
}

@misc{smoothquant,
      title={SmoothQuant: Accurate and Efficient Post-Training Quantization for Large Language Models}, 
      author={Guangxuan Xiao and Ji Lin and Mickael Seznec and Hao Wu and Julien Demouth and Song Han},
      year={2024},
      eprint={2211.10438},
      archivePrefix={arXiv},
      primaryClass={cs.CL},
      url={https://arxiv.org/abs/2211.10438}, 
}

@misc{gptq,
      title={GPTQ: Accurate Post-Training Quantization for Generative Pre-trained Transformers}, 
      author={Elias Frantar and Saleh Ashkboos and Torsten Hoefler and Dan Alistarh},
      year={2023},
      eprint={2210.17323},
      archivePrefix={arXiv},
      primaryClass={cs.LG},
      url={https://arxiv.org/abs/2210.17323}, 
}

@misc{awq,
      title={AWQ: Activation-aware Weight Quantization for LLM Compression and Acceleration}, 
      author={Ji Lin and Jiaming Tang and Haotian Tang and Shang Yang and Wei-Ming Chen and Wei-Chen Wang and Guangxuan Xiao and Xingyu Dang and Chuang Gan and Song Han},
      year={2024},
      eprint={2306.00978},
      archivePrefix={arXiv},
      primaryClass={cs.CL},
      url={https://arxiv.org/abs/2306.00978}, 
}

@misc{omniquant,
      title={OmniQuant: Omnidirectionally Calibrated Quantization for Large Language Models}, 
      author={Wenqi Shao and Mengzhao Chen and Zhaoyang Zhang and Peng Xu and Lirui Zhao and Zhiqian Li and Kaipeng Zhang and Peng Gao and Yu Qiao and Ping Luo},
      year={2024},
      eprint={2308.13137},
      archivePrefix={arXiv},
      primaryClass={cs.LG},
      url={https://arxiv.org/abs/2308.13137}, 
}

@misc{spqr,
      title={SpQR: A Sparse-Quantized Representation for Near-Lossless LLM Weight Compression}, 
      author={Tim Dettmers and Ruslan Svirschevski and Vage Egiazarian and Denis Kuznedelev and Elias Frantar and Saleh Ashkboos and Alexander Borzunov and Torsten Hoefler and Dan Alistarh},
      year={2023},
      eprint={2306.03078},
      archivePrefix={arXiv},
      primaryClass={cs.CL},
      url={https://arxiv.org/abs/2306.03078}, 
}

@misc{aqlm,
      title={Extreme Compression of Large Language Models via Additive Quantization}, 
      author={Vage Egiazarian and Andrei Panferov and Denis Kuznedelev and Elias Frantar and Artem Babenko and Dan Alistarh},
      year={2024},
      eprint={2401.06118},
      archivePrefix={arXiv},
      primaryClass={cs.LG},
      url={https://arxiv.org/abs/2401.06118}, 
}

@inproceedings{llmfp4,
   title={LLM-FP4: 4-Bit Floating-Point Quantized Transformers},
   url={http://dx.doi.org/10.18653/v1/2023.emnlp-main.39},
   DOI={10.18653/v1/2023.emnlp-main.39},
   booktitle={Proceedings of the 2023 Conference on Empirical Methods in Natural Language Processing},
   publisher={Association for Computational Linguistics},
   author={Liu, Shih-yang and Liu, Zechun and Huang, Xijie and Dong, Pingcheng and Cheng, Kwang-Ting},
   year={2023},
   pages={592–605} }

@misc{posttrainingquantizationdiffusionmodels,
      title={Post-training Quantization on Diffusion Models}, 
      author={Yuzhang Shang and Zhihang Yuan and Bin Xie and Bingzhe Wu and Yan Yan},
      year={2023},
      eprint={2211.15736},
      archivePrefix={arXiv},
      primaryClass={cs.CV},
      url={https://arxiv.org/abs/2211.15736}, 
}

@misc{qdiffusion,
      title={Q-Diffusion: Quantizing Diffusion Models}, 
      author={Xiuyu Li and Yijiang Liu and Long Lian and Huanrui Yang and Zhen Dong and Daniel Kang and Shanghang Zhang and Kurt Keutzer},
      year={2023},
      eprint={2302.04304},
      archivePrefix={arXiv},
      primaryClass={cs.CV},
      url={https://arxiv.org/abs/2302.04304}, 
}

@misc{ptqd,
      title={PTQD: Accurate Post-Training Quantization for Diffusion Models}, 
      author={Yefei He and Luping Liu and Jing Liu and Weijia Wu and Hong Zhou and Bohan Zhuang},
      year={2023},
      eprint={2305.10657},
      archivePrefix={arXiv},
      primaryClass={cs.CV},
      url={https://arxiv.org/abs/2305.10657}, 
}

@misc{svdquant,
      title={SVDQuant: Absorbing Outliers by Low-Rank Components for 4-Bit Diffusion Models}, 
      author={Muyang Li and Yujun Lin and Zhekai Zhang and Tianle Cai and Xiuyu Li and Junxian Guo and Enze Xie and Chenlin Meng and Jun-Yan Zhu and Song Han},
      year={2025},
      eprint={2411.05007},
      archivePrefix={arXiv},
      primaryClass={cs.CV},
      url={https://arxiv.org/abs/2411.05007}, 
}

@misc{treegrpo,
      title={TreeGRPO: Tree-Advantage GRPO for Online RL Post-Training of Diffusion Models}, 
      author={Zheng Ding and Weirui Ye},
      year={2025},
      eprint={2512.08153},
      archivePrefix={arXiv},
      primaryClass={cs.LG},
      url={https://arxiv.org/abs/2512.08153}, 
}

@misc{expandgrpo,
      title={Expand and Prune: Maximizing Trajectory Diversity for Effective GRPO in Generative Models}, 
      author={Shiran Ge and Chenyi Huang and Yuang Ai and Qihang Fan and Huaibo Huang and Ran He},
      year={2025},
      eprint={2512.15347},
      archivePrefix={arXiv},
      primaryClass={cs.CV},
      url={https://arxiv.org/abs/2512.15347}, 
}

@article{qerl,
      title={QeRL: Beyond Efficiency -- Quantization-enhanced Reinforcement Learning for LLMs}, 
      author={Wei Huang and Yi Ge and Shuai Yang and Yicheng Xiao and Huizi Mao and Yujun Lin and Hanrong Ye and Sifei Liu and Ka Chun Cheung and Hongxu Yin and Yao Lu and Xiaojuan Qi and Song Han and Yukang Chen},
      year={2025},
      eprint={2510.11696},
      archivePrefix={arXiv},
      primaryClass={cs.LG}
}

@misc{xi2026jetrlenablingonpolicyfp8,
      title={Jet-RL: Enabling On-Policy FP8 Reinforcement Learning with Unified Training and Rollout Precision Flow}, 
      author={Haocheng Xi and Charlie Ruan and Peiyuan Liao and Yujun Lin and Han Cai and Yilong Zhao and Shuo Yang and Kurt Keutzer and Song Han and Ligeng Zhu},
      year={2026},
      eprint={2601.14243},
      archivePrefix={arXiv},
      primaryClass={cs.LG},
      url={https://arxiv.org/abs/2601.14243}, 
}

@misc{flashrl_offpolicy,
  title = {Your Efficient RL Framework Secretly Brings You Off-Policy RL Training},
  url = {https://fengyao.notion.site/off-policy-rl},
  author = {Yao, Feng and Liu, Liyuan and Zhang, Dinghuai and Dong, Chengyu and Shang, Jingbo and Gao, Jianfeng},
  journal = {Feng Yao's Notion},
  year = {2025},
  month = aug,
}

@misc{flashrl,
  title = {FlashRL: 8Bit Rollouts, Full Power RL},
  url = {https://fengyao.notion.site/flash-rl},
  author = {Liu, Liyuan and Yao, Feng and Zhang, Dinghuai and Dong, Chengyu and Shang, Jingbo and Gao, Jianfeng},
  journal = {Feng Yao's Notion},
  year = {2025},
  month = aug,
}

@article{lee2023aligning,
  title={Aligning text-to-image models using human feedback},
  author={Lee, Kimin and Liu, Hao and Ryu, Moonkyung and Watkins, Olivia and Du, Yuqing and Boutilier, Craig and Abbeel, Pieter and Ghavamzadeh, Mohammad and Gu, Shixiang Shane},
  journal={arXiv preprint arXiv:2302.12192},
  year={2023}
}

@article{domingo2024adjoint,
  title={Adjoint matching: Fine-tuning flow and diffusion generative models with memoryless stochastic optimal control},
  author={Domingo-Enrich, Carles and Drozdzal, Michal and Karrer, Brian and Chen, Ricky TQ},
  journal={arXiv preprint arXiv:2409.08861},
  year={2024}
}

@article{black2023training,
  title={Training diffusion models with reinforcement learning},
  author={Black, Kevin and Janner, Michael and Du, Yilun and Kostrikov, Ilya and Levine, Sergey},
  journal={arXiv preprint arXiv:2305.13301},
  year={2023}
}

@article{liu2025flow,
  title={Flow-grpo: Training flow matching models via online rl},
  author={Liu, Jie and Liu, Gongye and Liang, Jiajun and Li, Yangguang and Liu, Jiaheng and Wang, Xintao and Wan, Pengfei and Zhang, Di and Ouyang, Wanli},
  journal={arXiv preprint arXiv:2505.05470},
  year={2025}
}

@article{xue2025dancegrpo,
  title={DanceGRPO: Unleashing GRPO on Visual Generation},
  author={Xue, Zeyue and Wu, Jie and Gao, Yu and Kong, Fangyuan and Zhu, Lingting and Chen, Mengzhao and Liu, Zhiheng and Liu, Wei and Guo, Qiushan and Huang, Weilin and others},
  journal={arXiv preprint arXiv:2505.07818},
  year={2025}
}

@article{shao2024deepseekmath,
  title={Deepseekmath: Pushing the limits of mathematical reasoning in open language models},
  author={Shao, Zhihong and Wang, Peiyi and Zhu, Qihao and Xu, Runxin and Song, Junxiao and Bi, Xiao and Zhang, Haowei and Zhang, Mingchuan and Li, YK and Wu, Yang and others},
  journal={arXiv preprint arXiv:2402.03300},
  year={2024}
}

@article{xu2023imagereward,
  title={Imagereward: Learning and evaluating human preferences for text-to-image generation},
  author={Xu, Jiazheng and Liu, Xiao and Wu, Yuchen and Tong, Yuxuan and Li, Qinkai and Ding, Ming and Tang, Jie and Dong, Yuxiao},
  journal={Advances in Neural Information Processing Systems},
  volume={36},
  pages={15903--15935},
  year={2023}
}

@article{clark2023directly,
  title={Directly fine-tuning diffusion models on differentiable rewards},
  author={Clark, Kevin and Vicol, Paul and Swersky, Kevin and Fleet, David J},
  journal={arXiv preprint arXiv:2309.17400},
  year={2023}
}

@book{pontryagin2018mathematical,
  title={Mathematical theory of optimal processes},
  author={Pontryagin, Lev Semenovich},
  year={2018},
  publisher={Routledge}
}

@article{kim2024pagoda,
  title={Pagoda: Progressive growing of a one-step generator from a low-resolution diffusion teacher},
  author={Kim, Dongjun and Lai, Chieh-Hsin and Liao, Wei-Hsiang and Takida, Yuhta and Murata, Naoki and Uesaka, Toshimitsu and Mitsufuji, Yuki and Ermon, Stefano},
  journal={Advances in Neural Information Processing Systems},
  volume={37},
  pages={19167--19208},
  year={2024}
}

@article{li2024reward,
  title={Reward guided latent consistency distillation},
  author={Li, Jiachen and Feng, Weixi and Chen, Wenhu and Wang, William Yang},
  journal={arXiv preprint arXiv:2403.11027},
  year={2024}
}

@article{luo2024diff,
  title={Diff-instruct*: Towards human-preferred one-step text-to-image generative models},
  author={Luo, Weijian and Zhang, Colin and Zhang, Debing and Geng, Zhengyang},
  journal={arXiv e-prints},
  pages={arXiv--2410},
  year={2024}
}

@article{luo2025reward,
  title={Reward-Instruct: A Reward-Centric Approach to Fast Photo-Realistic Image Generation},
  author={Luo, Yihong and Hu, Tianyang and Luo, Weijian and Kawaguchi, Kenji and Tang, Jing},
  journal={arXiv preprint arXiv:2503.13070},
  year={2025}
}

@article{fan2023dpok,
  title={Dpok: Reinforcement learning for fine-tuning text-to-image diffusion models},
  author={Fan, Ying and Watkins, Olivia and Du, Yuqing and Liu, Hao and Ryu, Moonkyung and Boutilier, Craig and Abbeel, Pieter and Ghavamzadeh, Mohammad and Lee, Kangwook and Lee, Kimin},
  journal={Advances in Neural Information Processing Systems},
  volume={36},
  pages={79858--79885},
  year={2023}
}

@article{schulman2017proximal,
  title={Proximal policy optimization algorithms},
  author={Schulman, John and Wolski, Filip and Dhariwal, Prafulla and Radford, Alec and Klimov, Oleg},
  journal={arXiv preprint arXiv:1707.06347},
  year={2017}
}

@inproceedings{fan2025online,
  title={Online reward-weighted fine-tuning of flow matching with wasserstein regularization},
  author={Fan, Jiajun and Shen, Shuaike and Cheng, Chaoran and Chen, Yuxin and Liang, Chumeng and Liu, Ge},
  booktitle={The Thirteenth International Conference on Learning Representations},
  year={2025}
}

@article{ouyang2022training,
  title={Training language models to follow instructions with human feedback},
  author={Ouyang, Long and Wu, Jeffrey and Jiang, Xu and Almeida, Diogo and Wainwright, Carroll and Mishkin, Pamela and Zhang, Chong and Agarwal, Sandhini and Slama, Katarina and Ray, Alex and others},
  journal={Advances in neural information processing systems},
  volume={35},
  pages={27730--27744},
  year={2022}
}

@article{bai2022training,
  title={Training a helpful and harmless assistant with reinforcement learning from human feedback},
  author={Bai, Yuntao and Jones, Andy and Ndousse, Kamal and Askell, Amanda and Chen, Anna and DasSarma, Nova and Drain, Dawn and Fort, Stanislav and Ganguli, Deep and Henighan, Tom and others},
  journal={arXiv preprint arXiv:2204.05862},
  year={2022}
}

@misc{deng2026densegrposparsedensereward,
      title={DenseGRPO: From Sparse to Dense Reward for Flow Matching Model Alignment}, 
      author={Haoyou Deng and Keyu Yan and Chaojie Mao and Xiang Wang and Yu Liu and Changxin Gao and Nong Sang},
      year={2026},
      eprint={2601.20218},
      archivePrefix={arXiv},
      primaryClass={cs.CV},
      url={https://arxiv.org/abs/2601.20218}, 
}

@article{guo2025deepseek,
  title={Deepseek-r1: Incentivizing reasoning capability in llms via reinforcement learning},
  author={Guo, Daya and Yang, Dejian and Zhang, Haowei and Song, Junxiao and Zhang, Ruoyu and Xu, Runxin and Zhu, Qihao and Ma, Shirong and Wang, Peiyi and Bi, Xiao and others},
  journal={arXiv preprint arXiv:2501.12948},
  year={2025}
}

@article{williams1992simple,
  title={Simple statistical gradient-following algorithms for connectionist reinforcement learning},
  author={Williams, Ronald J},
  journal={Machine learning},
  volume={8},
  number={3},
  pages={229--256},
  year={1992},
  publisher={Springer}
}

@misc{flux2024,
    author={Black Forest Labs},
    title={FLUX},
    year={2024},
    howpublished={\url{https://github.com/black-forest-labs/flux}},
}

@article{he2025tempflow,
  title={Tempflow-grpo: When timing matters for grpo in flow models},
  author={He, Xiaoxuan and Fu, Siming and Zhao, Yuke and Li, Wanli and Yang, Jian and Yin, Dacheng and Rao, Fengyun and Zhang, Bo},
  journal={arXiv preprint arXiv:2508.04324},
  year={2025}
}

@article{li2025mixgrpo,
  title={Mixgrpo: Unlocking flow-based grpo efficiency with mixed ode-sde},
  author={Li, Junzhe and Cui, Yutao and Huang, Tao and Ma, Yinping and Fan, Chun and Yang, Miles and Zhong, Zhao},
  journal={arXiv preprint arXiv:2507.21802},
  year={2025}
}

@article{li2025branchgrpo,
  title={BranchGRPO: Stable and Efficient GRPO with Structured Branching in Diffusion Models},
  author={Li, Yuming and Wang, Yikai and Zhu, Yuying and Zhao, Zhongyu and Lu, Ming and She, Qi and Zhang, Shanghang},
  journal={arXiv preprint arXiv:2509.06040},
  year={2025}
}

@article{mcallister2025flow,
  title={Flow Matching Policy Gradients},
  author={McAllister, David and Ge, Songwei and Yi, Brent and Kim, Chung Min and Weber, Ethan and Choi, Hongsuk and Feng, Haiwen and Kanazawa, Angjoo},
  journal={arXiv preprint arXiv:2507.21053},
  year={2025}
}

@article{greensmith2004variance,
  title={Variance reduction techniques for gradient estimates in reinforcement learning},
  author={Greensmith, Evan and Bartlett, Peter L and Baxter, Jonathan},
  journal={Journal of Machine Learning Research},
  volume={5},
  number={Nov},
  pages={1471--1530},
  year={2004}
}

@article{hu2022lora,
  title={Lora: Low-rank adaptation of large language models.},
  author={Hu, Edward J and Shen, Yelong and Wallis, Phillip and Allen-Zhu, Zeyuan and Li, Yuanzhi and Wang, Shean and Wang, Lu and Chen, Weizhu and others},
  journal={ICLR},
  volume={1},
  number={2},
  pages={3},
  year={2022}
}

@article{kirstain2023pick,
  title={Pick-a-pic: An open dataset of user preferences for text-to-image generation},
  author={Kirstain, Yuval and Polyak, Adam and Singer, Uriel and Matiana, Shahbuland and Penna, Joe and Levy, Omer},
  journal={Advances in neural information processing systems},
  volume={36},
  pages={36652--36663},
  year={2023}
}

@misc{hpsv2,
      title={Human Preference Score v2: A Solid Benchmark for Evaluating Human Preferences of Text-to-Image Synthesis}, 
      author={Xiaoshi Wu and Yiming Hao and Keqiang Sun and Yixiong Chen and Feng Zhu and Rui Zhao and Hongsheng Li},
      year={2023},
      eprint={2306.09341},
      archivePrefix={arXiv},
      primaryClass={cs.CV},
      url={https://arxiv.org/abs/2306.09341}, 
}

@article{chen2023textdiffuser,
  title={Textdiffuser: Diffusion models as text painters},
  author={Chen, Jingye and Huang, Yupan and Lv, Tengchao and Cui, Lei and Chen, Qifeng and Wei, Furu},
  journal={Advances in Neural Information Processing Systems},
  volume={36},
  pages={9353--9387},
  year={2023}
}

@article{hu2025brorl,
  title={Brorl: Scaling reinforcement learning via broadened exploration},
  author={Hu, Jian and Liu, Mingjie and Lu, Ximing and Wu, Fang and Harchaoui, Zaid and Diao, Shizhe and Choi, Yejin and Molchanov, Pavlo and Yang, Jun and Kautz, Jan and others},
  journal={arXiv preprint arXiv:2510.01180},
  year={2025}
}

@misc{clipscore,
      title={CLIPScore: A Reference-free Evaluation Metric for Image Captioning}, 
      author={Jack Hessel and Ari Holtzman and Maxwell Forbes and Ronan Le Bras and Yejin Choi},
      year={2022},
      eprint={2104.08718},
      archivePrefix={arXiv},
      primaryClass={cs.CV},
      url={https://arxiv.org/abs/2104.08718}, 
}

@misc{xie2024sana,
      title={Sana: Efficient High-Resolution Image Synthesis with Linear Diffusion Transformer}, 
      author={Enze Xie and Junsong Chen and Junyu Chen and Han Cai and Haotian Tang and Yujun Lin and Zhekai Zhang and Muyang Li and Ligeng Zhu and Yao Lu and Song Han},
      year={2024},
      eprint={2410.10629},
      archivePrefix={arXiv},
      primaryClass={cs.CV}
}

@article{sd,
      title={Scaling Rectified Flow Transformers for High-Resolution Image Synthesis},
      author={Esser, Patrick and Kulal, Sumith and Blattmann, Andreas and Entezari, Rahim and M{\"u}ller, Jonas and Saini, Harry and Levi, Yam and Lorenz, Dominik and Sauer, Axel and Boesel, Frederic and others},
      journal={arXiv preprint arXiv:2403.03206},
      year={2024}
}

@misc{transformerengine,
      title={Transformer Engine: A library for accelerating Transformer models on NVIDIA GPUs}, 
      author={{NVIDIA}},
      year={2022},
      howpublished={\url{https://github.com/NVIDIA/TransformerEngine}}
}

@misc{qurl,
      title={QuRL: Efficient Reinforcement Learning with Quantized Rollout}, 
      author={Yuhang Li and Reena Elangovan and Xin Dong and Priyadarshini Panda and Brucek Khailany},
      year={2026},
      eprint={2602.13953},
      archivePrefix={arXiv},
      primaryClass={cs.LG},
      url={https://arxiv.org/abs/2602.13953}, 
}

@misc{vespo,
      title={VESPO: Variational Sequence-Level Soft Policy Optimization for Stable Off-Policy LLM Training}, 
      author={Guobin Shen and Chenxiao Zhao and Xiang Cheng and Lei Huang and Xing Yu},
      year={2026},
      eprint={2602.10693},
      archivePrefix={arXiv},
      primaryClass={cs.LG},
      url={https://arxiv.org/abs/2602.10693}, 
}

@misc{fp8-rl,
      title={FP8-RL: A Practical and Stable Low-Precision Stack for LLM Reinforcement Learning}, 
      author={Zhaopeng Qiu and Shuang Yu and Jingqi Zhang and Shuai Zhang and Xue Huang and Jingyi Yang and Junjie Lai},
      year={2026},
      eprint={2601.18150},
      archivePrefix={arXiv},
      primaryClass={cs.LG},
      url={https://arxiv.org/abs/2601.18150}, 
}

@misc{fu2025dynamictreerpobreakingindependenttrajectory,
      title={Dynamic-TreeRPO: Breaking the Independent Trajectory Bottleneck with Structured Sampling}, 
      author={Xiaolong Fu and Lichen Ma and Zipeng Guo and Gaojing Zhou and Chongxiao Wang and ShiPing Dong and Shizhe Zhou and Shizhe Zhou and Ximan Liu and Jingling Fu and Tan Lit Sin and Yu Shi and Zhen Chen and Junshi Huang and Jason Li},
      year={2025},
      eprint={2509.23352},
      archivePrefix={arXiv},
      primaryClass={cs.CV},
      url={https://arxiv.org/abs/2509.23352}, 
}

@article{chen2025bridging,
  title={Bridging supervised learning and reinforcement learning in math reasoning},
  author={Chen, Huayu and Zheng, Kaiwen and Zhang, Qinsheng and Cui, Ganqu and Cui, Yin and Ye, Haotian and Lin, Tsung-Yi and Liu, Ming-Yu and Zhu, Jun and Wang, Haoxiang},
  journal={arXiv preprint arXiv:2505.18116},
  year={2025}
}

@article{chen2025towards,
  title={Towards self-improvement of diffusion models via group preference optimization},
  author={Chen, Renjie and Lin, Wenfeng and Zhang, Yichen and Wei, Jiangchuan and Liu, Boyuan and Feng, Chao and Ran, Jiao and Guo, Mingyu},
  journal={arXiv preprint arXiv:2505.11070},
  year={2025}
}

@article{luo2025reinforcing,
  title={Reinforcing Diffusion Models by Direct Group Preference Optimization},
  author={Luo, Yihong and Hu, Tianyang and Tang, Jing},
  journal={arXiv preprint arXiv:2510.08425},
  year={2025}
}

@article{choi2026rethinking,
  title={Rethinking the Design Space of Reinforcement Learning for Diffusion Models: On the Importance of Likelihood Estimation Beyond Loss Design},
  author={Choi, Jaemoo and Zhu, Yuchen and Guo, Wei and Molodyk, Petr and Yuan, Bo and Bai, Jinbin and Xin, Yi and Tao, Molei and Chen, Yongxin},
  journal={arXiv preprint arXiv:2602.04663},
  year={2026}
}

@inproceedings{wallace2024diffusion,
  title={Diffusion model alignment using direct preference optimization},
  author={Wallace, Bram and Dang, Meihua and Rafailov, Rafael and Zhou, Linqi and Lou, Aaron and Purushwalkam, Senthil and Ermon, Stefano and Xiong, Caiming and Joty, Shafiq and Naik, Nikhil},
  booktitle={Proceedings of the IEEE/CVF Conference on Computer Vision and Pattern Recognition},
  pages={8228--8238},
  year={2024}
}
}

\end{document}